%% file: main.tex
\documentclass{article}

\usepackage{amsthm}

\usepackage[accent=0072B2,font=palatino,headings=sans]{simpleicml}
\usepackage{sectionnav}
\usepackage{capt-of} 

\usepackage{graphicx} 
\usepackage{wrapfig}
\usepackage{booktabs}
\usepackage{multirow}
\usepackage{makecell}
\usepackage{pifont}
\usepackage{colortbl}
\usepackage{xcolor}
\usepackage{array}
\usepackage{caption}
\usepackage{enumitem}

\definecolor{headerblue}{HTML}{EBF3FF}
\definecolor{textred}{HTML}{980000}
\definecolor{textgreen}{HTML}{274E13}
\definecolor{mygray}{gray}{0.5}

\newcommand{\negval}[1]{\textcolor{textred}{\scriptsize (#1)}}
\newcommand{\posval}[1]{\textcolor{textgreen}{\scriptsize (#1)}}

\definecolor{lightblue}{RGB}{220,230,241}
\definecolor{graytext}{gray}{0.4}

\definecolor{darkblue}{rgb}{0.0,0.0,0.7}
\definecolor{darkred}{rgb}{0.7,0.0,0.0}
\newcommand{\cmark}{\textcolor{darkblue}{\ding{51}}}
\newcommand{\xmark}{\textcolor{darkred}{\ding{55}}}

\usepackage{hyperref} 
\hypersetup{
    colorlinks=true,
    linkcolor=black,
    citecolor=darkblue,
    urlcolor=darkblue,
}

\icmltitle{VL-JEPA: Joint Embedding Predictive Architecture for Vision-language}

\icmlauthors{%
  Delong Chen\textsuperscript{1,2}*, \quad
  Mustafa Shukor\textsuperscript{1,3}*, \quad
  Théo Moutakanni\textsuperscript{1}*, \quad
  Willy Chung\textsuperscript{1,3}*, \quad \\
  Jade Yu\textsuperscript{1}, 
  Tejaswi Kasarla\textsuperscript{1}, 
  Yejin Bang\textsuperscript{1}, 
  Allen Bolourchi\textsuperscript{1}, 
  Yann LeCun\textsuperscript{1,4}, 
  Pascale Fung\textsuperscript{1,2}
}
\icmlaffiliations{%
  \textsuperscript{1} Meta FAIR \quad 
  \textsuperscript{2} HKUST \quad
  \textsuperscript{3} Sorbonne Université \quad
  \textsuperscript{4} NYU \\
  * Equal contribution\\
  \texttt{delong.chen@connect.ust.hk}
}
\icmlabstract{
\input{content/abstract}
}
\icmlrunningtitle{VL-JEPA: Joint Embedding Predictive Architecture for Vision-language}
\input{math_commands}

\sectionheaderline{
  \seclink{sec:intro}{1}{Intro} | 
  \seclink{sec:method}{2}{Method} | 
  \seclink{sec:implementation-details}{3}{Implementation} |
  \seclink{sec:experiments}{4}{Experiments} |
  \seclink{sec:related-works}{5}{Related Works} |
  \seclink{sec:conclusion}{6}{Conclusion}
}

\begin{document}
\icmlmaketitle

\input{document_commands}

\input{content/intro}

\input{content/method}

\input{content/implementation}

\input{content/experiments}

\input{content/related}
\input{content/conclusion}

\bibliography{bibliography}
\bibliographystyle{plainnat}

\end{document}

%% file: content/abstract.tex
We introduce VL-JEPA, a vision-language model built on a Joint Embedding Predictive Architecture (JEPA). Instead of autoregressively generating tokens as in classical VLMs, VL-JEPA predicts \textit{continuous embeddings} of the target texts. By learning in an abstract representation space, the model focuses on task-relevant semantics while abstracting away surface-level linguistic variability. In a strictly controlled comparison against standard token-space VLM training with the same vision encoder and training data, VL-JEPA achieves stronger performance while having 50\% fewer trainable parameters. At inference time, a lightweight text decoder is invoked only when needed to translate VL-JEPA predicted embeddings into text. We show that VL-JEPA natively supports \textit{selective decoding} that reduces the number of decoding operations by $\sim$2.85$\times$ while maintaining similar performance compared to non-adaptive uniform decoding. Beyond generation, the VL-JEPA's embedding space naturally supports open-vocabulary classification, text-to-video retrieval, and discriminative VQA \textit{without any architecture modification}. On eight video classification and eight video retrieval datasets, the average performance VL-JEPA surpasses that of CLIP, SigLIP2, and Perception Encoder. At the same time, the model achieves comparable performance as classical VLMs (InstructBLIP, QwenVL) on four VQA datasets: GQA, TallyQA, POPE and POPEv2, despite only having 1.6B parameters.

%% file: math_commands.tex
\usepackage{natbib}
\usepackage{hyperref}
\usepackage{microtype}
\usepackage{graphicx}
\usepackage{subfigure}
\usepackage{booktabs} 
\usepackage[toc,page,header]{appendix}
\usepackage{minitoc}
\usepackage{makecell}
\usepackage{dblfloatfix}
\usepackage{amsmath}
\usepackage[capitalize,noabbrev]{cleveref}
\usepackage{mathtools}
\usepackage{cuted} 
\usepackage[most]{tcolorbox}
\tcbuselibrary{theorems}
\usepackage{caption}
\usepackage{enumitem}
\usepackage{listings}

\usepackage{tgheros} 
\definecolor{bg}{HTML}{F8FAFD}
\definecolor{keyword}{HTML}{0077B6}
\definecolor{string}{HTML}{E76F51}
\definecolor{comment}{HTML}{A0AEC0}
\definecolor{number}{HTML}{457B9D}
\definecolor{function}{HTML}{2A9D8F}
\definecolor{class}{HTML}{F4A261}
\definecolor{text}{HTML}{2D3A4A}
\lstdefinestyle{pytorchheros}{
    backgroundcolor=\color{bg},
    basicstyle=\fontfamily{qhv}\selectfont\footnotesize\color{text}, 
    keywordstyle=\color{keyword}\bfseries,
    stringstyle=\color{string},
    commentstyle=\color{comment}\itshape,
    numberstyle=\color{number},
    identifierstyle=\color{text},
    classoffset=1,
    morekeywords={Net},
    keywordstyle=\color{class}\bfseries,
    classoffset=0,
    emph={__init__,forward,print}, emphstyle=\color{function},
    frame=single,
    framerule=0pt,
    rulecolor=\color{bg},
    tabsize=4,
    showstringspaces=false,
    breaklines=true,
    linewidth=\linewidth,
    xleftmargin=0em,
    xrightmargin=0em,
    aboveskip=1em,
    belowskip=1em,
    literate={~}{{\textasciitilde}}1
}
\renewcommand\lstlistingname{Algorithm}
\crefname{lstlisting}{\MakeLowercase\lstlistingname}{\MakeLowercase\lstlistingname s}
\Crefname{lstlisting}{\lstlistingname}{\lstlistingname s}
\usepackage{float} 
\newfloat{lstfloat}{htbp}{lop}
\floatname{lstfloat}{Listing}

\usepackage{amsmath,amsthm}
\usepackage{bm,bbm}

\usepackage{varwidth}

\usepackage{hyperref}
\usepackage{cleveref}
\usepackage{mathtools}
\usepackage{algpseudocode}
\usepackage{mdframed}

\usepackage{multicol}
\usepackage{multirow}
\usepackage{setspace}

\usepackage{colortbl}

\usepackage[svgnames]{xcolor}
\usepackage{framed}

\newcommand{\1}{\mathbf{1}}

\usepackage{tikz,ifthen}
\usetikzlibrary{positioning}
\usetikzlibrary{shapes}
\usetikzlibrary{arrows}
\usetikzlibrary{fit}
\usetikzlibrary{calc}
\usetikzlibrary{shapes.misc}

\crefname{defi}{defn.}{defns.}

\def\1{\bm{1}}

\DeclareMathAlphabet{\mathsfit}{\encodingdefault}{\sfdefault}{m}{sl}
\SetMathAlphabet{\mathsfit}{bold}{\encodingdefault}{\sfdefault}{bx}{n}



%% file: document_commands.tex
\newtcbtheorem
  [
  crefname={detail}{detail}]
  {detail}
  {Experiment Details}
  {%
    fontupper=\small,
    colback=orange!5,
    colframe=orange!35!black,
    fonttitle=\bfseries,
    boxsep=1pt,
    left=1.5mm,
    right=1.5mm,
    top=2mm,
    bottom=1mm,
  }
  {detail}

\newtcbtheorem
  [
  crefname={def.}{def.}]
  {definition}
  {Definition}
  {%
    fontupper=\small,
    colback=green!5,
    colframe=green!35!black,
    fonttitle=\bfseries,
    boxsep=1pt,
    left=1.5mm,
    right=1.5mm,
    top=2mm,
    bottom=1mm,
  }
  {def}

\newtcbtheorem
  [
  crefname={thm.}{thms.}]
  {theorem}
  {Theorem}
  {%
    fontupper=\small,
    colback=red!5,
    colframe=red!35!black,
    fonttitle=\bfseries,
    boxsep=1pt,
    left=1.5mm,
    right=1.5mm,
    top=2mm,
    bottom=1mm,
  }
  {theorem}
\newtcbtheorem
  [
  crefname={lemma.}{lemmas.}]
  {lemma}
  {Lemma}
  {%
    fontupper=\small,
    colback=blue!3,
    colframe=blue!35!black,
    fonttitle=\bfseries,
    boxsep=1pt,
    left=1.5mm,
    right=1.5mm,
    top=2mm,
    bottom=1mm,
  }
  {lemma}
\newtcbtheorem
  [
  crefname={prop.}{props.}]
  {proposition}
  {Proposition}
  {%
    breakable,
    enhanced,
    fontupper=\small,
    colback=red!5,
    colframe=red!35!black,
    fonttitle=\bfseries,
    boxsep=1pt,
    left=1.5mm,
    right=1.5mm,
    top=2mm,
    bottom=1mm,
  }
  {proposition}
\newtcbtheorem
  [
  crefname={cor.}{corrs.}]
  {corollary}
  {Corollary}
  {%
    breakable,
    enhanced,
    fontupper=\small,
    colback=red!5,
    colframe=red!35!black,
    fonttitle=\bfseries,
    boxsep=1pt,
    left=1.5mm,
    right=1.5mm,
    top=2mm,
    bottom=1mm,
  }
  {corollary}

%% file: content/intro.tex
\medskip
\section{Introduction}
\label{sec:intro}

One of the most important aspects of advanced machine intelligence is the ability to understand the physical world that surrounds us. This ability enables AI systems to learn, reason, plan and act in the real world in order to assist humans \citep{lecun2022path}. Intelligent systems that need to act in the real world includes wearable devices and robots \citep{fung2025embodied}. Machine learning tasks that make up for this ability include captioning, retrieval, visual question answering, action tracking, reasoning and planning etc \citep{bordes2024introduction, chen2025planning}. Systems for such real-world applications must have real-time response with low latency and inference cost.

Currently, the common approach to achieve these tasks is to use large token-generative Vision Language Models (VLMs) \citep{liu2023visual, dai2023instructblip, alayrac2022flamingo, chen2024visual, cho2025perceptionlm, chen2022pali}, which takes visual input $X_V$, textual query $X_Q$ to generate desired textual response $Y$ autoregressively in token space, \textit{i.e.,} $(X_V, X_Q) \mapsto Y$. This is straightforward but inadequate for two main reasons. First, VLMs are expensive to develop, because they are trained to generate responses $Y$ to queries by capturing both task-relevant semantics with task-irrelevant surface linguistic features such as words choice, style or paraphrasing. During training, VLMs must model both aspects, which results in unnecessary computing effort spent producing diverse token sequences that ultimately do not impact the correctness of the output. Second, real-time tasks involving live streaming video (\textit{e.g.,} live action tracking) require sparse and selective decoding (\textit{e.g.,}, emitting a description only when a new event occurs) \citep{zhou2024streaming}. However, VLMs rely on autoregressive token-by-token decoding, which must be completed before revealing the underlying semantics of $Y$. This process introduces unnecessary latency and hampers the ability to update semantics dynamically in real time.

\begin{figure}[t!]
    \centering
    \vspace{+20pt}
    \includegraphics[width=0.55\linewidth]{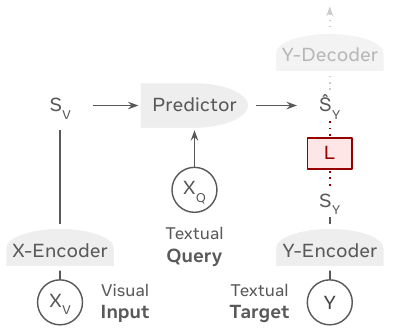}
    \caption{\centering{VL-JEPA model architecture}}
    \label{fig:vl_jepa_small}
    \vspace{-15pt}
\end{figure}

This paper introduces the Joint Embedding Predictive Architecture for Vision-Language (VL-JEPA), turning expensive learning of data-space token generation into more efficient latent-space semantic prediction. As illustrated in Fig.~\ref{fig:vl_jepa_small}, the model employs \textbf{x-encoder} to map vision inputs $X_V$ into embedding $S_V$, a \textbf{y-encoder} to map the textual target $Y$ into an embedding $S_Y$, and a \textbf{predictor} that learns the mapping $(S_V, X_Q) \mapsto S_Y$ where $X_Q$ is a textual query (\textit{i.e.,} the prompt). The training objective is defined in the embedding space $\mathcal{L}_\texttt{VL-JEPA} = D(\hat{S}_Y, S_Y)$ instead of the data space $\mathcal{L}_\texttt{VLM} = D(\hat{Y}, Y)$. During inference, a \textbf{y-decoder} reads out the predicted embedding $\hat{S}_Y$ to text space $\hat{Y}$ when needed.

Thanks to its \textbf{non-generative} nature, VL-JEPA is not forced to reconstruct every surface detail of $Y$ in the token space. Instead, it only needs to predict the abstract representation $S_Y$ in the embedding space. In the raw one-hot token space, different plausible $Y$ outputs for the same input often appear nearly orthogonal if they don't share overlapping tokens. However, in the embedding space, these diverse targets can be mapped to nearby points that share similar semantics. This simplifies the target distribution thus makes the learning process more efficient. In addition, unlike VLMs, this approach eliminates the need for learning language generation with a heavy decoder during training, resulting in significant efficiency gains.

Thanks to its \textbf{non-autoregressive} nature, VL-JEPA can produce continuous streams of target semantic embeddings within sliding windows with minimal latency as it only require a single forward pass without autoregressive decoding. This is particularly advantageous for real-time online applications such as live action tracking, scene recognition, or planning, where the embedding stream can be selectively decoded by a lightweight y-decoder, enabling efficient and prompt updates.

In this work, we empirically validate the advantages of VL-JEPA. We conduct a strictly controlled comparison against classical token-generative VLM \citep{liu2023visual, cho2025perceptionlm}: both setups use the same vision encoder, spatial resolution, frame rate, training data, batch size, and number of iterations, etc., with the \emph{only} difference being the objective in token space or embedding space. Under this matched training condition, VL-JEPA delivers consistently higher performance on zero-shot captioning and classification while using roughly half the trainable parameters, indicating that embedding-space supervision improves learning efficiency. 

Beyond the training phase, VL-JEPA also delivers substantial inference-time efficiency improvement through \emph{selective decoding}, where decoding happens only due to significant change in the predicted embedding stream. Empirically, this strategy reduces the number of decoding operations by $\sim$2.85$\times$ while preserving overall output quality measured by average CIDEr scores.

{Our final VL-JEPA models are trained in two stages: 1) a pretraining stage using caption data to establish robust vision-language alignment, and 2) a supervised finetuning (SFT) stage that equips the model with VQA capabilities. The model resulting from the first stage, denoted as \textbf{VL-JEPA$_\texttt{BASE}$},} is evaluated on \textit{zero-shot} classification and text-to-video retrieval. VL-JEPA$_\texttt{BASE}$ outperforms CLIP \citep{radford2021learning}, SigLIP2 \citep{tschannen2025siglip}, and Perception Encoder \citep{bolya2025perception} models in terms of average classification accuracy (across 8 datasets) and retrieval recall@1 (across 8 datasets). {Following the second stage, the resulting \textbf{VL-JEPA$_\texttt{SFT}$} demonstrates significantly improved classification performance due to its exposure to in-domain training data. As a unified \textit{generalist} model, VL-JEPA$_\texttt{SFT}$ approaches the performance of \textit{specialist} models optimized for individual benchmarks. Simultaneously, VL-JEPA$_\texttt{SFT}$ exhibits effective VQA capabilities, achieving performance on par with established VLM families}, such as InstructBLIP \citep{dai2023instructblip} and Qwen-VL \citep{bai2023qwen}, across four datasets covering compositional visual reasoning \citep{hudson2019gqa}, complex object counting \citep{acharya2019tallyqa}, and object hallucination \citep{li2023evaluating, li2025analyzing}. 

In summary, the contributions of this paper are as follows:

\begin{itemize}

    \item We introduce VL-JEPA, the first non-generative model that can perform general-domain vision-language tasks in real-time, built on a joint embedding predictive architecture. 
    
    \item We demonstrate in controlled experiments that VL-JEPA, trained with latent space embedding prediction, outperforms VLMs that rely on data space token prediction.
    
    \item We show that VL-JEPA delivers significant efficiency gains over VLMs for online video streaming applications, thanks to its non-autoregressive design and native support for selective decoding.
    
    \item {We highlight that our VL-JEPA$_\texttt{SFT}$ model, with an unified model architecture, can effectively handle a wide range of classification, retrieval, and VQA tasks at the same time.}

\end{itemize}

%% file: content/method.tex
\begin{figure*}
    \centering
    \includegraphics[width=\linewidth]{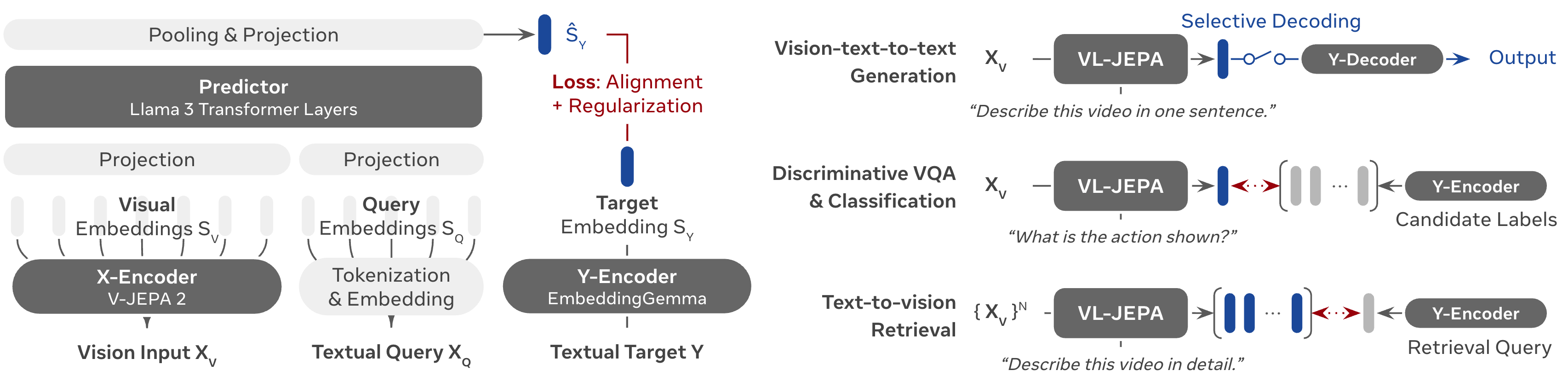}
    \caption{
        \textbf{Left: VL-JEPA Architecture}. It learns to predict the target embedding $S_Y$, instead of reconstructing the raw target $Y$ in token space as in classical VLMs. 
        \textbf{Right: VL-JEPA Applications}: It handles vision-text-to-text generation tasks (\textit{e.g.,} captioning) with selective decoding mechanism natively supported. Furthermore, VL-JEPA's embedding space facilitates {discriminative VQA}, open-vocabulary classification and text-to-video retrieval tasks using a single unified model architecture.
    }
    \label{fig:vl_jepa}
\end{figure*}

\section{Methodology}
\label{sec:method}

We propose \textbf{VL-JEPA} (Fig.~\ref{fig:vl_jepa_small}), a model with the joint embedding predictive architecture (JEPA) for vision-language tasks. VL-JEPA is trained with triplets $\langle X_V, X_Q, Y \rangle$, where $X_V$ denotes the \textbf{visual input} (a single image or a sequence of video frames), $X_Q$ is a \textbf{textual query} (\textit{i.e.,} a question) and $Y$ is the \textbf{textual target} (\textit{i.e.,} the answer) to be predicted. The VL-JEPA comprises of four components:

\begin{enumerate}[leftmargin=3em, itemsep=2pt]
    \item \textbf{\texttt{X-Encoder}} $(X_V \mapsto S_V)$ compresses high-volume visual inputs to compact visual embeddings--a sequence of continuous vectors analogous to ``visual tokens'' in classical VLMs.

    \item \textbf{\texttt{Predictor}} $(\langle S_V, X_Q\rangle \mapsto \hat{S}_Y)$ is the core component of VL-JEPA. It maps visual embeddings to a prediction of target embedding, with a textual query as conditioning. 

    \item \textbf{\texttt{Y-Encoder}} $(Y \mapsto S_Y)$ embeds the textual target into a continuous latent space as the prediction target. The target embedding is expected to abstract away task irrelevant information. 
    
    \item \textbf{\texttt{Y-Decoder}} $(\hat{S}_Y \mapsto \hat{Y})$ is not involved during the main training phrase of VL-JEPA. At inference time, it translates the predicted embedding as human-readable text when necessary. 
\end{enumerate}


Fig.~\ref{fig:vl_jepa} illustrates how we instantiate the VL-JEPA architecture in this paper. For the \texttt{X-Encoder}, we chose V-JEPA~2 \citep{assran2025v}, a Vision Transformer that outputs a sequence of visual tokens, which are then projected and fed into the \texttt{Predictor} initialized using Llama 3 Transformer layers. Query conditioning is achieved by tokenizing and embedding the textual query and feeding the resulting textual token embeddings into the \texttt{Predictor} along with the visual embeddings. The outputs of the Llama 3 Transformer layers are pooled and projected into the target embedding space produced by the \texttt{Y-Encoder}, {which is initialized by EmbeddingGemma-300M}~\citep{vera2025embeddinggemma}. We provide more technical details in \S\ref{sec:implementation-details}.
 
\textbf{Training Objective.} JEPA models typically optimize two objectives jointly: 1) prediction error in the embedding space, and 2) additional regularization that avoids representation collapse \citep{bardes2021vicreg, balestriero2025lejepa}. Any loss that implements these two properties can be applied to VL-JEPA. Alternatively, the regularization term can be replaced by other anti-collapse strategies, such as using an exponential moving average (EMA) for the \texttt{Y-Encoder} \citep{assran2025v} or freezing the \texttt{Y-Encoder} \citep{zhou2025dino}.

In this work, {we adopt the \textbf{InfoNCE loss} \citep{radford2021learning} due to its maturity in the vision-language domain.} More advanced non-sample-contrastive regularization, such as VICReg \citep{bardes2021vicreg} and SIGReg \citep{balestriero2025lejepa} can also be applied but we leave the exploration to future works. {InfoNCE loss can be mathematically divided \citep{wang2020understanding} into: 1) a representation \textit{alignment} term that minimizes the distance between normalized prediction and target embeddings, and 2) a \textit{uniformity} regularization term that pushes embeddings in a batch apart from each other, thus avoiding representation collapse. We train the \texttt{Predictor} and the \texttt{Y-Encoder} jointly with bi-directional InfoNCE loss, enabling them to mutually learn from each other.}

Compared to the token-space loss used by generative VLMs, calculating the training loss in the embedding space is beneficial due to the \textbf{simplified target distribution}. Specifically, many real-world prediction tasks are inherently ill-posed: for the same input $X$, there may exist multiple plausible targets $Y$ that are all acceptable. For example, given the query \textit{``What will happen here if I flip this light switch down?''}, both \textit{``the lamp is turned off''} and \textit{``room will go dark''} are valid answers. In the raw one-hot token space, however, the two sequences are orthogonal since they share no overlapping tokens. But when VL-JEPA’s \texttt{Y-Encoder} embeds them into nearby points (ideally yielding a compact unimodal distribution), the learning task becomes much easier: the model no longer needs to fit multiple disjoint high-density regions in sparse token space, but only a single coherent mode in a continuous embedding space.

\textbf{Multi-tasking.} VL-JEPA supports diverse tasks using a \textit{single}, \textit{unified} architecture (Fig.~\ref{fig:vl_jepa}). For vision-text-to-text generation tasks, such as captioning or open-ended VQA, the query $X_Q$ is a captioning prompt or a question, and the predictor learns to predict the embedding of the target output, $\hat S_Y$, which is then decoded into text. {VL-JEPA also supports CLIP-style open-vocabulary classification and discriminative VQA}, where candidate label texts are encoded into embeddings and compared with prediction $\hat S_Y$ to select the nearest match. For text-to-video retrieval, candidate videos are mapped to their predicted embeddings $\hat S_Y$ using a retrieval a captioning prompt, and then ranked by similarity to the encoded textual retrieval query.

\textbf{Selective Decoding.} Real-world video applications often require online streaming inference, such as tracking user actions in smart glasses for procedural assistance \citep{chen2024videollm}, monitoring world states for online planning, navigation and robotics \citep{shukor2025smolvla, black2025real,song2025accelerating}. A central challenge is balancing two competing needs: the model must continuously update semantics as new frames arrive, but  computational efficiency and latency are critical. 

Existing VLMs typically rely on explicit memory mechanisms \citep{zhou2024streaming,qian2024streaming} to decide when to decode or complex KV-cache optimizations \citep{di2025streaming} for efficiency, since autoregressive language models are expensive to run continuously. 
VL-JEPA, in contrast, natively supports selective decoding. Since it predicts a semantic answer embedding non-autoregressively, the model provides a continuous semantic stream of $\hat S_Y$ that can be monitored in real time. This stream can be stabilized with simple smoothing (\textit{e.g.,} average pooling) and decoded only when a significant semantic shift is detected, such as when the local window variance exceeds a threshold. In this way, VL-JEPA maintains always-on semantic monitoring while avoiding unnecessary decoding, achieving both responsiveness and efficiency.

%% file: content/implementation.tex
\section{Implementation of VL-JEPA}
\label{sec:implementation-details}

\subsection{Model Architecture}

\textbf{{X-Encoder}.} Unless otherwise specified, we use a frozen \texttt{V-JEPA~2 ViT-L} \citep{assran2025v} with 304M parameters, a self-supervised vision model that excels at both image and video tasks. Each video input is uniformly sampled into frames at 256$^2$ resolution. For image inputs, the same image is duplicated to match the input shape.

\textbf{{Predictor}.} The predictor is initialized with the last 8 Transformer layers of \texttt{Llama-3.2-1B}, resulting in 490M trainable parameters. The text tokenizer and token embedding are also from \texttt{Llama-3.2-1B}. We allow maximum 512 query tokens, and put \texttt{[PAD]} tokens for short queries. We disable the causal attention mask so that both vision and query embeddings can be jointly attended. Linear projections connect the predictor with the vision and text embeddings, and average pooling on non-\texttt{[PAD]} tokens is applied to obtain the predicted target embedding.

\textbf{{Y-Encoder}.} {We use \texttt{EmbeddingGemma-300M}~\citep{vera2025embeddinggemma} as the initialization of the \texttt{Y-Encoder}. We set maximum context length of 512 to handle detailed captions. We found that setting a learning rate multiplier of $\times$0.05 to all text encoder parameters improves performance, since the quality of embedding prediction would be suboptimal in the beginning of training.} Linear projection head is applied to both \texttt{Predictor} and \texttt{Y-Encoder}, obtaining a shared embedding space with 1,536 dimensions, where the loss is calculated.

{\subsection{Two-stage Training}}

\textbf{Large-scale Pretraining.} VL-JEPA is trained with two stages. The first query-free pretraining stage aims to establish robust vision-language alignment using massive caption data. We use Datacomp \citep{gadre2023datacomp} and YFCC-100M \citep{thomee2016yfcc100m} for image-text data. For video-text data, we use \textsc{Action100M} \citep{chen2026action100m}, which consists action description and video captions generated on HowTo100M videos \citep{chen2025planning}.

We first do image-only training on Datacomp and YFCC-100M with only 1 frame per visual input, which allows us to use a large batch size of 24k. After 100k iterations, the model has seen 2B samples and achieved 61.6\% ImageNet zero-shot accuracy (without prompt ensembling). Then, we continue with video pretraining with 8 frames per input for 60k iterations and 32 frames for 10k iterations in the end. The pretraining takes 4 weeks using 24 nodes with 8$\times$NVIDIA H200 GPUs each. We adopt a constant learning rate of $5{\times}10^{-5}$ to facilitate extended training. We refer the readers to the Action100M paper \citep{chen2026action100m} for more details. We call the resulting model \textbf{VL-JEPA$_\texttt{BASE}$} and measure \textit{zero-shot} classification and retrieval performance with this model.

\textbf{Supervised Finetuning.} The second query-conditioned supervised finetuning (SFT) stage empowers VL-JEPA VQA capabilities while maintaining the pretrained vision-language alignment for classification and retrieval. The training data is selected from the PLM data mixture \citep{cho2025perceptionlm}, including 25M VQA samples, 2.8M captioning samples, 1.8M classification samples, and downsampled pretraining stage data to avoid catastrophic forgetting.

We train the model for 83k steps with a batch size of 3,072 ($\sim$2.5s days with 24 nodes), with cosine learning rate annealing applied to improve convergence. Since excessive human labeled data is included in this SFT data mixture, we no longer emphasize \textit{zero-shot} evaluation for the resulting \textbf{VL-JEPA$_\texttt{SFT}$} from this stage. Instead, we evaluate VQA capabilities and compare it with state-of-the-art \textit{specialist} models.

%% file: content/experiments.tex
\begin{table*}[h!]
\caption{\textbf{Video classification and text-to-video retrieval}. Best \textit{zero-shot} performance in each dataset are \underline{\textbf{highlighted}}. Samples seen $=$ training step $\times$ effective batch size. 
}
\label{tab:cls_ret}
\resizebox{\linewidth}{!}{%
\begin{tabular}{ccrrcc|c|cccccccc|c|cccccccc}
\multicolumn{2}{c}{} & & & & \multicolumn{1}{l|}{} & \multicolumn{9}{c|}{\textbf{Video Classification} (Top-1 Accuracy)} & \multicolumn{9}{c}{\textbf{Text-to-video Retrieval} (Recall@1)} \\
\multicolumn{2}{c}{\multirow{-2}{*}{\textbf{Model}}} & \rotatebox{90}{\textbf{\# Parameters}} & \rotatebox{90}{\textbf{\# Samples Seen}} & \rotatebox{90}{\textbf{Zero-shot}} & \rotatebox{90}{\textbf{Generalist Model}} & \rotatebox{90}{\textbf{Average}} & 
\rotatebox[origin=bl]{90}{\shortstack[l]{SSv2\\{\scriptsize\citep{goyal2017something}}}} &
\rotatebox[origin=bl]{90}{\shortstack[l]{EK100\\{\scriptsize\citep{damen2022rescaling}}}} &
\rotatebox[origin=bl]{90}{\shortstack[l]{EgoExo4D\\{\scriptsize\citep{grauman2024ego}}}} &
\rotatebox[origin=bl]{90}{\shortstack[l]{Kinetics-400\\{\scriptsize\citep{kay2017kinetics}}}} &
\rotatebox[origin=bl]{90}{\shortstack[l]{COIN (SR)\\{\scriptsize\citep{tang2019coin}}}} &
\rotatebox[origin=bl]{90}{\shortstack[l]{COIN (TR)\\{\scriptsize\citep{tang2019coin}}}} &
\rotatebox[origin=bl]{90}{\shortstack[l]{CrossTask (SR)\\{\scriptsize\citep{zhukov2019cross}}}} &
\rotatebox[origin=bl]{90}{\shortstack[l]{CrossTask (TR)\\{\scriptsize\citep{zhukov2019cross}}}} &
\rotatebox[origin=bl]{90}{\shortstack[l]{\textbf{Average}}} &
\rotatebox[origin=bl]{90}{\shortstack[l]{MSR-VTT\\{\scriptsize\citep{xu2016msr}}}} &
\rotatebox[origin=bl]{90}{\shortstack[l]{ActivityNet\\{\scriptsize\citep{caba2015activitynet}}}} &
\rotatebox[origin=bl]{90}{\shortstack[l]{DiDeMo\\{\scriptsize\citep{anne2017localizing}}}} &
\rotatebox[origin=bl]{90}{\shortstack[l]{MSVD\\{\scriptsize\citep{chen2011collecting}}}} &
\rotatebox[origin=bl]{90}{\shortstack[l]{YouCook2\\{\scriptsize\citep{zhou2018towards}}}} &
\rotatebox[origin=bl]{90}{\shortstack[l]{PVD-Bench\\{\scriptsize\citep{bolya2025perception}}}} &
\rotatebox[origin=bl]{90}{\shortstack[l]{Dream-1k\\{\scriptsize\citep{wang2024tarsier}}}} &
\rotatebox[origin=bl]{90}{\shortstack[l]{VDC-1k\\{\scriptsize\citep{chai2024auroracap}}}}
\\ \toprule
 & RN50 & 75M & 12.8B &  & & 21.8 & 2.1 & 1.5 & 2.1 & 41.4 & 8.6 & 39.0 & 10.9 & 68.7 & 28.3 & 28.7 & 17.7 & 24.7 & 29.7 & 5.1 & 27.6 & 47.2 & 46.0 \\
 & ViT-B & 124M & 12.8B &  &  & 25.4 & 3.1 & 1.3 & 2.8 & 49.5 & 11.2 & 47.3 & 16.2 & 71.5 & 29.3 & 31.0 & 19.5 & 25.7 & 34.0 & 6.1 & 27.0 & 48.5 & 42.9 \\
\multirow{-3}{*}{CLIP} & ViT-L & 389M & 12.8B & \multirow{-3}{*}{\cmark} & \multirow{-3}{*}{\cmark} & 30.7 & 3.8 & 3.7 & 2.6 & 58.3 & 14.7 & 63.5 & 20.8 & 78.5 & 35.3 & 35.9 & 23.4 & 30.7 & 41.9 & 7.9 & 36.7 & 56.8 & 49.3 \\ \midrule
 & ViT-B & 375M & 40B &  &  & 33.9 & 5.2 & 2.3 & 4.5 & 57.8 & 20.6 & 69.9 & 27.7 & 82.9 & 39.6 & 40.2 & 25.0 & 32.1 & 48.6 & 13.8 & 52.1 & 60.9 & 43.7 \\
 & ViT-L & 882M & 40B &  &  & 38.6 & 5.9 & 4.5 & 6.4 & 63.6 & 24.2 & 78.5 & 35.1 & 90.8 & 45.4 & 41.6 & 32.7 & 35.1 & 53.5 & 19.0 & 59.2 & 71.6 & 50.9 \\
\multirow{-3}{*}{SigLIP2} & ViT-g & 1.9B & 40B & \multirow{-3}{*}{\cmark} & \multirow{-3}{*}{\cmark} & 39.8 & 6.1 & 6.1 & 5.6 & 68.0 & 26.0 & 80.4 & 35.1 & 90.8 & 47.5 & 43.4 & 33.9 & 38.9 & 56.0 & 22.2 & 60.4 & 73.0 & 52.5 \\ \midrule
 & ViT-B & 448M & 58B &  &  & 37.2 & 5.8 & 3.3 & 6.0 & 65.4 & 21.5 & 77.1 & 26.9 & 91.8 & 44.9 & 46.5 & 35.4 & 35.3 & 49.1 & 15.2 & 59.8 & 68.7 & 49.2 \\
 & ViT-L & 671M & 58B &  &  & 42.9 & 9.3 & 6.0 & 11.6 & 73.4 & 27.1 & 83.3 & 37.5 & 95.3 & 50.2 & 48.9 & 41.7 & 40.8 & 56.2 & 22.5 & 64.7 & 75.9 & 51.0 \\
\multirow{-3}{*}{PE-Core} & ViT-G & 2.3B & 86B & \multirow{-3}{*}{\cmark} & \multirow{-3}{*}{\cmark} & 44.7 & 9.0 & 6.4 & 13.6 & \underline{\textbf{76.4}} & 29.0 & \underline{\textbf{86.0}} & 40.3 & \underline{\textbf{97.2}} & 58.1 & \underline{\textbf{51.6}} & 49.1 & 44.5 & \underline{\textbf{58.7}} & 26.0 & 77.0 & 89.2 & 68.5 \\ \midrule
\rowcolor[HTML]{EBF3FF} 
\textbf{VL-JEPA$_\texttt{BASE}$} & ViT-L & 1.6B & 3.3B & \cmark & \cmark & \underline{\textbf{52.5}} & \underline{\textbf{19.3}} & \underline{\textbf{21.8}} & \underline{\textbf{33.2}} & 64.8 & \underline{\textbf{47.4}} & 79.4 & \underline{\textbf{64.5}} & 89.6 & \underline{\textbf{63.7}} & 40.0 & \underline{\textbf{64.9}} & \underline{\textbf{50.0}} & 49.0 & \underline{\textbf{40.4}} & \underline{\textbf{83.1}} & \underline{\textbf{93.3}} & \underline{\textbf{88.8}} \\ \bottomrule \toprule
\rowcolor[HTML]{EBF3FF} 
\textbf{VL-JEPA$_\texttt{SFT}$} & ViT-L & 1.6B & 3.6B & \xmark & \cmark & \textit{75.4} & \textit{73.2} & \textit{44.6} & \textit{68.1} & \textit{84.8} & \textit{66.4} & \textit{90.3} & \textit{79.8} & \textit{96.2} & \textit{63.8} & \textit{46.2} & \textit{62.4} & \textit{52.3} & \textit{52.3} & \textit{37.8} & \textit{82.6} & \textit{89.3} & \textit{87.7} \\ \midrule
\multicolumn{4}{c}{\textit{\color{mygray} Previous SoTA (including specialist models)}} & \textit{\color{mygray} \xmark} & \textit{\color{mygray} \xmark} & - & \textit{\color{mygray} 77.5} & \textit{\color{mygray} 56.4} & \textit{\color{mygray} 47.8} & \textit{\color{mygray} 92.1} & \textit{\color{mygray} 67.3} & \textit{\color{mygray} 95.3} & \textit{\color{mygray} 64.5} & \textit{\color{mygray} 96.0} & - & \textit{\color{mygray} 62.8} & \textit{\color{mygray} 74.1} & \textit{\color{mygray} 74.2} & \textit{\color{mygray} 61.4} & \textit{\color{mygray} 28.9} & \textit{\color{mygray} 77.0} & \textit{\color{mygray} 89.2} & \textit{\color{mygray} 68.5} \\ \bottomrule
\end{tabular}%
}
\end{table*}

\section{Experiments}
\label{sec:experiments}

\looseness=-1
We begin by evaluating VL-JEPA's classification and retrieval performance in \S\ref{sec:exp.cls_ret}, and {benchmark VL-JEPA on VQA datasets in \S\ref{sec:exp.vqa}.} We demonstrate application of VL-JEPA for understanding the relationship between world state changes and action concepts (\textit{i.e.,} inverse dynamics) in \S\ref{sec:exp.worldprediction}. In \S\ref{sec:exp.embbeddings_vs_tokens}, we demonstrate the advantage of embedding prediction by comparing it with a token-predictive VLM baseline under a strictly controlled setting. In \S\ref{sec:exp.selective decoding}, we evaluate the effectiveness of VL-JEPA's selective decoding, and show that it reduces decoding cost while maintaining the performance. Next, we analyze VL-JEPA's \texttt{Y-Encoder} in \S\ref{sec:exp.y-encoder}. Finally, we present ablation studies in \S\ref{sec:exp.ablations}. 

\subsection{Classification and Retrieval}
\label{sec:exp.cls_ret}

\textbf{Evaluation Setup.}  We evaluate VL-JEPA following the CLIP-style evaluation protocol (see Fig.\ref{fig:vl_jepa} and \S\ref{sec:method} ``Multi-tasking''). We assess VL-JEPA on a broad suite of benchmarks, including 8 classification datasets and 8 retrieval datasets. For \textit{zero-shot} evaluation, we compare against \textit{generalist foundation models} CLIP~\citep{radford2021learning}, SigLIP2~\citep{tschannen2025siglip}, and Perception Encoder (PE-Core)\citep{bolya2025perception}. We additionally report reference numbers from \textit{specialist models} that are individually optimized for each benchmark.

\textbf{Results.} Table~\ref{tab:cls_ret} summarizes the results. In the strict zero-shot setting, VL-JEPA$_\texttt{BASE}$ achieves higher average accuracy (52.5 vs 44.7) across the 8 classification datasets and higher average recall@1 (63.7 vs 58.1) across the 8 retrieval datasets than the best baseline PE-Core-G. Per-dataset scores show that VL-JEPA$_\texttt{BASE}$ is {particularly strong on \textit{motion-centric} benchmarks (SSv2, EK-100, EgoExo4D, and step recognition on COIN and CrossTask), while relatively weaker on \textit{appearance-centric} benchmarks (Kinetics-400 and task recognition on COIN and CrossTask)}. This is due to VL-JEPA$_\texttt{BASE}$ has seen substantially fewer vision-language pairs (only 3.6B in comparison with PE-Core-G's 86B). After supervised finetuning, VL-JEPA$_\texttt{SFT}$ improves significantly upon VL-JEPA$_\texttt{BASE}$ since the model has seen in-domain training data. As a single \textit{generalist} model, the performance of VL-JEPA$_\texttt{SFT}$ is approaching \textit{specialist} models optimized individually for each dataset.

\begin{table*}[h!]
\caption{{\textbf{VQA benchmarks.} We report accuracy on GQA \citep{hudson2019gqa}, TallyQA \citep{acharya2019tallyqa}, POPE \citep{li2023evaluating}, and POPEv2 \citep{li2025analyzing}. Scores lower than our model are marked in \textcolor[HTML]{980000}{red}. Scores from SmolVLM  are obtained by our evaluation, while other baselines are reported in the literature.}}
\label{tab:holistic_benchmarks}
\centering

\resizebox{\textwidth}{!}{%
\begin{tabular}{lr c lr c lr c lr}

\multicolumn{2}{c}{\textit{\textbf{GQA}: compositional visual reasoning}} & & 
\multicolumn{2}{c}{\textit{\textbf{TallyQA}: complex object counting}} & & 
\multicolumn{2}{c}{\textit{\textbf{POPE}: object hallucination}} & & 
\multicolumn{2}{c}{\textit{\textbf{POPEv2}: object hallucination}} \\

\cmidrule{1-2} \cmidrule{4-5} \cmidrule{7-8} \cmidrule{10-11}

\textbf{Model} & \textbf{Accuracy} & & 
\textbf{Model} & \textbf{Accuracy} & & 
\textbf{Model} & \textbf{Accuracy} & & 
\textbf{Model} & \textbf{Accuracy} \\

\cmidrule{1-2} \cmidrule{4-5} \cmidrule{7-8} \cmidrule{10-11}


BLIP-2 (OPT-2.7B) & {\color[HTML]{980000} 33.9} & & 
SmolVLM-256M & {\color[HTML]{980000} 32.3} & & 
SmolVLM2-256M & {\color[HTML]{980000} 56.4} & & 
SmolVLM-256M & {\color[HTML]{980000} 62.3} \\

BLIP-2 (FlanT5XXL) & {\color[HTML]{980000} 41.0} & & 
SmolVLM-500M & {\color[HTML]{980000} 44.8} & & 
SmolVLM-256M & {\color[HTML]{980000} 57.9} & & 
LLaVA-1.5-13B & {\color[HTML]{980000} 72.7} \\

InstructBLIP (FlanT5XL) & {\color[HTML]{980000} 48.4} & & 
PaLI-700M & {\color[HTML]{980000} 62.3} & & 
LLaVA-7B & {\color[HTML]{980000} 72.9} & & 
InternVL2-8B & {\color[HTML]{980000} 74.5} \\

InstructBLIP (Vicuna-13B) & {\color[HTML]{980000} 49.5} & & 
SmolVLM-2B & {\color[HTML]{980000} 64.7} & & 
InstructBLIP (Vicuna-13B) & {\color[HTML]{980000} 79.0} & & 
InternVL2-26B & {\color[HTML]{980000} 76.1} \\

Qwen-VL-Chat-7B & {\color[HTML]{980000} 57.5} & & 
PaLI-3B & {\color[HTML]{980000} 65.8} & & 
Video-LLaVA (7B) & {\color[HTML]{980000} 83.4} & & 
Qwen2-VL-72B & {\color[HTML]{980000} 79.4} \\

Qwen-VL-7B & {\color[HTML]{980000} 59.3} & & 
InstructBLIP (Vicuna-13B) & {\color[HTML]{980000} 68.0} & & 
SmolVLM-500M & 85.8 & & 
SmolVLM-500M & {\color[HTML]{980000} 83.8} \\

InternVL-Chat (Vicuna-7B) & {\color[HTML]{980000} 59.5} & & 
PaLI-17B & 71.9 & & 
LLaVA-1.5-7B & 85.9 & & 
Qwen2-VL-7B & 87.0 \\

LLaVA-1.5 (Vicuna-7B) & 62.0 & & 
LLaVA-1.5 (Vicuna-13B) & 72.3 & & 
LLaVA-1.5-13B-HD & 86.3 & & 
SmolVLM-2B & 88.8 \\

InternVL-Chat (Vicuna-13B) & 66.6 & & 
PaliGemma (3B) & 76.8 & & 
SmolVLM-2B & 87.5 & & 
Qwen2-VL-2B & 91.3 \\

\cmidrule{1-2} \cmidrule{4-5} \cmidrule{7-8} \cmidrule{10-11}

\cellcolor[HTML]{EBF3FF}\textbf{VL-JEPA$_\texttt{SFT}$} (1.6B) & \cellcolor[HTML]{EBF3FF}61.5 & & 
\cellcolor[HTML]{EBF3FF}\textbf{VL-JEPA$_\texttt{SFT}$} (1.6B) & \cellcolor[HTML]{EBF3FF}69.9 & & 
\cellcolor[HTML]{EBF3FF}\textbf{VL-JEPA$_\texttt{SFT}$} (1.6B) & \cellcolor[HTML]{EBF3FF}85.7 & & 
\cellcolor[HTML]{EBF3FF}\textbf{VL-JEPA$_\texttt{SFT}$} (1.6B) & \cellcolor[HTML]{EBF3FF}86.3 \\

\cmidrule{1-2} \cmidrule{4-5} \cmidrule{7-8} \cmidrule{10-11}

\end{tabular}%
}
\end{table*}

\begin{table*}[h!]
\centering
\caption{\textbf{\textsc{WorldPrediction-WM} benchmark results}. We compare the accuracy between large VLMs, socratic LLMs, and VL-JEPA. VL-JEPA$_\texttt{SFT}$ achieves a new SoTA at 65.7\%.}
\label{tab:worldprediction}
\resizebox{\textwidth}{!}{%
\setlength{\tabcolsep}{5pt}
\renewcommand{\arraystretch}{1.2}
\begin{tabular}{*{20}{c}}
\toprule
\multicolumn{8}{c}{\textbf{Vision Language Models}} &
\multicolumn{10}{c}{\textbf{Socratic LLMs} (w/ Qwen2.5-VL-72B captions)} &
\multicolumn{2}{c}{\textbf{VL-JEPA}} \\
\cmidrule(lr){1-8}\cmidrule(lr){9-18}\cmidrule(lr){19-20}
\multicolumn{4}{c}{\texttt{InternVL2.5}} &
\multicolumn{4}{c}{\texttt{Qwen2.5-VL}} &
\multicolumn{2}{c}{\texttt{Llama-3.1}} &
\multicolumn{2}{c}{\texttt{Llama-4}} &
\multicolumn{3}{c}{\texttt{Qwen2.5}} &
\multicolumn{1}{c}{\texttt{GPT-4o}} &
\multicolumn{1}{c}{\texttt{Claude-3.5}} &
\multicolumn{1}{c}{\texttt{Gemini-2}} &
\multicolumn{1}{c}{\texttt{BASE}} &
\multicolumn{1}{c}{\texttt{SFT}} \\
\texttt{2B} & \texttt{4B} & \texttt{26B} & \texttt{38B} &
\texttt{3B} & \texttt{7B} & \texttt{32B} & \texttt{72B} &
\texttt{8B} & \texttt{70B} &
\texttt{109B} & \texttt{400B} &
\texttt{3B} & \texttt{7B} & \texttt{72B} &
\texttt{N/A} & \texttt{N/A} & \texttt{N/A} &
\texttt{1.6B}&
\texttt{1.6B} \\
\midrule
20.0 & 29.8 & 30.2 & 50.3 &
21.6 & 45.5 & 49.0 & 57.0 &
48.7 & 49.8 &
52.7 & 53.6 &
44.0 & 49.1 & 48.5 &
52.0 & 53.3 & 55.6 &
\cellcolor{headerblue}63.9  &
\cellcolor{headerblue}\underline{\textbf{65.7}} \\
\bottomrule
\end{tabular}}
\end{table*}

\subsection{Visual Question Answering}
\label{sec:exp.vqa}

\textbf{Evaluation Setup.} We evaluate VL-JEPA$_\texttt{SFT}$ on discriminative VQA tasks. The inference process involves encode candidate answers using the \texttt{Y-Encoder} and selecting the answer that minimizes the distance to the predicted embedding (see Fig.~\ref{fig:vl_jepa}). We select four benchmarks that prioritize visual perception rather than knowledge and reasoning. We evaluate on GQA \citep{hudson2019gqa}, a dataset for real-world visual reasoning and compositional QA, reporting accuracy on the testdev-balanced split. For TallyQA \citep{acharya2019tallyqa}, which targets complex counting, we follow \cite{chen2022pali} and report the weighted average accuracy across the ``simple'' and ``complex'' splits. Finally, to assess object hallucination, we utilize POPE \citep{li2023evaluating} and POPEv2 \citep{li2025analyzing}. For POPE, we report the average accuracy across the ``random'', ``popular'', and ``adversarial'' settings on MS-COCO.

\textbf{Results.} Table~\ref{sec:exp.vqa} compares VL-JEPA$_\texttt{SFT}$ against established VLM families, including BLIP-2 \citep{li2023blip}, InstructBLIP \citep{dai2023instructblip}, Qwen-VL \citep{bai2023qwen}, InternVL \citep{chen2024internvl}, Llava-1.5 \citep{vallaeys2024improved}, SmolVLM \citep{marafioti2025smolvlm}, PaLI \citep{chen2022pali}, PaliGemma \citep{beyer2024paligemma}, and Video-LLaVA \citep{lin2024video}. VL-JEPA$_\texttt{SFT}$ outperforms many of these baselines despite requiring significantly less computational resources--classical VLMs rely on extensively pretrained CLIP backbones combined with multi-stage visual instruction tuning. In comparison, VL-JEPA$_\texttt{SFT}$ employs a \textit{unified architecture} and a \textit{single embedding space} to seamlessly handle VQA, classification, and retrieval (Tab.~\ref{tab:cls_ret}).

\subsection{WorldPrediction-WM} 
\label{sec:exp.worldprediction}

\textbf{Evaluation Setup.} We evaluate VL-JEPA on the ``world modeling’’ task in the \textsc{WorldPrediction}~\citep{chen2025worldprediction} benchmark, where the model is provided with two images representing the initial and final world states and must identify, among four candidate video clips, the action that explains the observed transition. To adapt VL-JEPA, we duplicate and concatenate the initial and final state images to extract a \textit{state embedding}, and encode each action candidate into \textit{action embeddings}. The model then selects the candidate whose embedding is closest to the state embedding.

\textbf{Results.} Table~\ref{tab:worldprediction} shows accuracy comparisons. {VL-JEPA$_\texttt{BASE}$ attains \textbf{63.9\%} and VL-JEPA$_\texttt{SFT}$ attains \textbf{65.7\%}} top-1 accuracy on \textsc{WorldPrediction}-WM, establishing a new state of the art. Our VL-JEPA model not only substantially surpasses existing VLMs of comparable or larger scale but also exceeds the performance of frontier LLMs such as GPT-4o, Claude-3.5-sonnet, and Gemini-2.0. 

\subsection{Action Anticipation}

\begin{table}
    \centering
    \caption{Finetuning results on EPIC-KITCHENS-100 action anticipation (Recall@5) for different anticipation time $t$. VL-JEPA outperforms V-JEPA2 with a ViT-L-256px vision encoder across all anticipation time.}
    \resizebox{\linewidth}{!}{
    \begin{tabular}{cccccc}
    \toprule
        \textbf{Model} & \textbf{Vision Encoder} & $t=1s$ & $t=2s$ & $t=4s$ & $t=10s$ \\\midrule
        
        {\color{mygray}V-JEPA2} & {\color{mygray}\texttt{ViT-g-384px}}  &{\color{mygray}\textbf{\underline{39.7}}}  & {\color{mygray}\textbf{\underline{28.6}}}  & {\color{mygray}19.3}  & {\color{mygray}8.2} \\
        V-JEPA2 & \texttt{ViT-L-256px} &  32.7 & 23.4 & 15.9 & 7.1\\
        \rowcolor{headerblue}
        VL-JEPA & \texttt{ViT-L-256px} & 34.2 & 26.0 & \textbf{\underline{19.4}} & \textbf{\underline{11.7}}\\\bottomrule
    \end{tabular}}
    \vspace{-10pt}
    \label{tab:action_anticipation}
\end{table}

\begin{table}[]
\centering
    \caption{Finetuning result on COIN next step forecasting. VL-JEPA established a new SoTA at 56.2.}
 \resizebox{\linewidth}{!}{%
    \begin{tabular}{ccccc}
    \toprule
    \textbf{Model} & \begin{tabular}[c]{@{}c@{}}VideoLLM\\ -online (8B)\end{tabular} & \begin{tabular}[c]{@{}c@{}}VideoLLM\\ -MoD (8B)\end{tabular} & \begin{tabular}[c]{@{}c@{}}ProVideLLM\\ -8B/11+\end{tabular} & \begin{tabular}[c]{@{}c@{}}VL-JEPA\\ (1.6B)\end{tabular}\\ 
        \midrule
    \begin{tabular}[c]{@{}c@{}}\textbf{Accuracy} \end{tabular} & 49.1 & 49.7 & 53.6 & \cellcolor{headerblue}\textbf{56.2} \\ \bottomrule
    \end{tabular}
}
\vspace{-10pt}
\label{tab:coin_forecasting}
\end{table}

\begin{figure*}
    \centering
    \includegraphics[width=\linewidth]{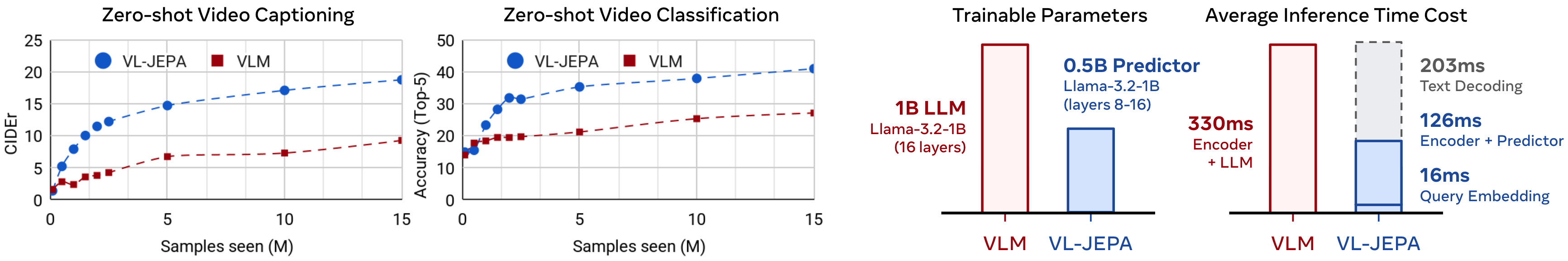}
    \caption{\textbf{Comparison of embedding prediction (VL-JEPA) and token prediction (VLM).} We conduct a fair comparison of under strictly aligned training settings (encoder, data, batchsize, etc.). \textbf{Left:} Zero-shot video captioning CIDEr score averaged over 3 datasets and zero-shot classification accuracy (top-5) averaged over 3 benchmarks. \textbf{Right:} Comparing the trainable parameters and average inference time cost.}
    \label{fig:vljepa_vs_vlm}
\end{figure*}

\textbf{Evauation Setup.} We assess the action anticipation (\textit{i.e.,} forecasting future actions) capabilities of fine-tuned VL-JEPA on two benchmarks: action anticipation with EPIC-KITCHENS-100 \citep{damen2022rescaling}, and COIN \citep{tang2019coin} next step forecasting. 
In this task, given a window of context video, the model is required to predict the next action that will occur after a specified anticipation time—the temporal gap between the end of the context segment and the onset of the subsequent action. 
VL-JEPA$_\texttt{SFT}$ is fine-tuned and we compare our results against the state-of-the-art baselines. 

\textbf{Results} As compated in Table VL-JEPA demonstrates strong performance in both benchmarks (Table~\ref{tab:action_anticipation} and Table~\ref{tab:coin_forecasting}). For the Epic-Kitchens-100, particularly at the standard 1-second anticipation interval, where it achieves a Recall@5 of 34.18 -- surpassing the previous V-JEPA 2 that uses the same ViT-L-256px encoder by 1.48 points. While the larger V-JEPA 2 ViT-g-384px model attains the highest score at 1 second, VL-JEPA remains competitive and shows notable advantages at longer anticipation intervals. 
On the COIN next step forecasting task, as shown in Tab. \ref{tab:coin_forecasting}, VL-JEPA achieves 56.2\%, outperform strong VLM baselines VideoLLM-online \citep{chen2024videollm}, VideoLLM-MoD \citep{wu2024videollm}, ProVideLLM \citep{chatterjee2025memory}. Overall, these findings demonstrated that VL-JEPA can be finetuned to handle semantic uncertain prediction tasks.

\subsection{Embedding Prediction vs. Token Prediction: A Controlled Comparison}
\label{sec:exp.embbeddings_vs_tokens}

\textbf{Evaluation Setup.} In this section, we compare VL-JEPA to a token-generative VLM baseline under a strictly aligned training conditions. Both models use the same Perception Encoder \citep{bolya2025perception} (frozen ViT-L-14 with 336$^2$ resolution, no tiling, 16 frames per video) for vision inputs. We use the same training iterations with the same effective batch size of 128, same learning rate scheduler on the same pretraining data mixture described above (\S\ref{sec:implementation-details}). The only difference is the prediction task: VL-JEPA predicts target embeddings \citep{duquenne2023sonar} using a 0.5B predictor, whereas the VLM baseline performs next-token prediction with cross-entropy using a 1B LLM. For VLM, we use the standard training recipe and codebase of PerceptionLM \citep{cho2025perceptionlm}, aligning frozen vision encoder and text-only LLM \texttt{Llama-3.2-1B}. For VL-JEPA, we initialize the predictor from the 8-16 layers of \texttt{Llama-3.2-1B}.

We evaluate both models at regular checkpoints throughout training spanning from 500K to 15M samples seen. At each checkpoint, we measure the performance on video captioning and video classification. For video captioning, we report CIDEr scores averaged across YouCook2 \citep{zhou2018towards}, MSR-VTT \citep{xu2016msr} and PVD-Bench \citep{bolya2025perception}. VL-JEPA decodes the predicted embeddings while VLM generates the tokens directly. For video classification, we report top-5 accuracy averaged across CrossTask-Step, CrossTask-Task \citep{zhukov2019cross} and EgoExo4D \citep{grauman2024ego}. For VL-JEPA we choose the candidate with lowest cosine distance to the predicted embedding, while for VLM we pick the class with lowest perplexity.

\textbf{Results.} As shown in Fig.~\ref{fig:vljepa_vs_vlm}, both models yield comparable performance after 500K samples seen in both tasks, with respectively 1.23 and 1.35 CIDEr in video captioning and 14.9\% and 14.0\% top-5 accuracy for VL-JEPA and VLM. After a few iterations, we show that VL-JEPA's performance increase is much sharper compared to VLM, reaching 14.7 CIDEr and 35.3\% top-5 accuracy after 5M samples seen. This gap remains constant as training scales at 15M samples with 14.8 CIDEr and 41.0\% top-5 accuracy for VL-JEPA, while the VLM baseline yield respectively 7.1 CIDEr and 27.2\% top-5 accuracy. This controlled comparison highlights the benefit of predicting embeddings rather than tokens, showing both higher sample efficiency and stronger absolute performance.

We compare the inference cost of the above VL-JEPA and the VLM by pre-loading 64 video frames into memory and repeatedly decoding text 100 times with the same prompt, measuring the average time per sample. As shown in Fig.~\ref{fig:vljepa_vs_vlm} (right most), both models exhibit comparable latency when generating text. What differentiates our model from classical VLM is the decoupling between the prompt processing (``Query Embedding'') and the video encoder (``Encoder + Predictor'') from the text generation module (``Decoder''). This allows us to only use the first part of the model to perform retrieval and decode text only when needed (see Section~\ref{sec:exp.selective decoding} below), making our model more scalable for online video inference.

\subsection{Effectiveness of Selective Decoding}
\label{sec:exp.selective decoding}

\begin{figure*}
    \centering
    \includegraphics[width=1\linewidth]{figures/selective_decoding.png}
    \caption{
        \textbf{Evaluation of selective decoding.}
        \textbf{Left:} 
            We compare uniform sampling of decoding points at fixed intervals (red) and embedding-guided selective decoding (blue). Performance is measured by the average CIDEr score between each annotation $y$ and its closest decoded output $\hat{y}$. 
        \textbf{Right:} 
            Results on EgoExo4D show that selective decoding achieves a Pareto improvement over uniform sampling: for the same performance level, it requires fewer decoding operations.
    }
    \label{fig:selective_decoding}
\end{figure*}

\begin{table*}[t!]
\centering
\caption{{\textbf{Comparison of text-encoders performance}. We report triplet-based accuracy (\%) on SugarCrepe++ and VISLA datasets.}}
\label{tab:y-decoder}
\resizebox{\linewidth}{!}{%
\begin{tabular}{llr r|c| ccccc|c| cc}
\multicolumn{2}{c}{\multirow{3}{*}{\textbf{Model}}} & \multirow{3}{*}{\makecell{\textbf{\# Params.}\\(total)}} & \multirow{3}{*}{\makecell{\textbf{\# Params.}\\(text encoder)}} & \multicolumn{6}{c|}{\textbf{SugarCrepe++} \citep{10.5555/3737916.3738487}} & \multicolumn{3}{c}{\textbf{VISLA} \citep{dumpala2024vislabenchmarkevaluatingembedding}} \\
 & & & & \multirow{2}{*}{\textbf{Average}} & \makecell{Replace\\Attribute} & \makecell{Replace\\Object} & \makecell{Replace\\Relation} & \makecell{Swap\\Attribute} & \makecell{Swap\\Object} & \multirow{2}{*}{\textbf{Average}} & \multirow{2}{*}{Generic} & \multirow{2}{*}{Spatial} \\
\toprule
CLIP & ViT-L & 389M & 85M & 44.5 & 56.7 & 83.0 & 42.5 & 27.0 & 13.5 & 34.5 & 37.6 & 31.3 \\
SigLIP2 & ViT-g & 1.9B & 708M & 56.5 & 66.9 & 74.4 & \underline{52.1} & \underline{58.4} & \underline{30.6} & \underline{40.4} & \underline{48.7} & 32.1 \\
PE-Core & ViT-G & 2.3B & 537M & \underline{58.6} & \textbf{73.6} & 90.6 & 48.9 & 53.2 & 26.5 & 38.3 & 45.2 & 31.4 \\
\midrule
\rowcolor[HTML]{EBF3FF}
\textbf{VL-JEPA$_\texttt{BASE}$} & ViT-L & 1.6B & 300M & \textbf{63.9} & \underline{72.2} & \underline{90.1} & \textbf{52.2} & \textbf{62.9} & \textbf{42.0} & \textbf{42.9} & \textbf{49.8} & \textbf{35.9} \\
\rowcolor[HTML]{EBF3FF}
\textbf{VL-JEPA$_\texttt{SFT}$} & ViT-L & 1.6B & 300M & 58.4 & 68.5 & \textbf{90.9} & 47.4 & 55.4 & 29.8 & 39.5 & 44.8 & \underline{34.2} \\
\bottomrule
\end{tabular}
}
\end{table*}

\looseness=-1
\textbf{Evaluation Setup.} We evaluate the effectiveness of VL-JEPA’s embedding-guided selective decoding on long-form video streams.  To this end, we design a benchmark task where the goal is to recover a temporal sequence of annotations while minimizing the number of text decoding operations, which dominate inference cost. As shown in Fig.~\ref{fig:selective_decoding} (left), decoding is performed only at selected points along the VL-JEPA embedding stream, yielding a sequence of $N$ decoded outputs $[(\hat{t}_1, \hat{y}_1), (\hat{t}_2, \hat{y}_2), \ldots, (\hat{t}_N, \hat{y}_N)]$. Each ground-truth annotation $[(t_1, y_1), (t_2, y_2), \ldots, (t_T, y_T)]$ is then aligned to its nearest decoded output in time (illustrated as $\circ \cdot \cdot \cdot \circ$ in Fig.~\ref{fig:selective_decoding}), and CIDEr is computed between matched pairs. We use the EgoExo4D \citep{grauman2024ego} validation set in procedural activity domains, which consists of 218 videos with an average duration of 6 minutes and about $T=143$ atomic action annotations per video.

As a baseline, we consider \textit{uniform sampling}, where decoding points are placed at fixed intervals regardless of the underlying video content. Standard streaming VLMs are limited to this strategy, whereas VL-JEPA supports a more effective alternative: \textit{adaptive selection} of decoding points guided by its predicted embeddings. We apply agglomerative clustering with temporal connectivity constraints \citep{murtagh2012algorithms} to partition the embedding sequence into $N$ segments of high intra-segment monosemanticity \citep{chen2024subobject}, measured by variance (\textit{i.e.,} Ward distance). The intuition is that within a semantically coherent segment, decoded outputs are highly similar, so decoding once per segment captures the essential information while greatly reducing overall decoding cost. The midpoint of each segment is then chosen as the decoding point, and decoding is performed either from the exact embedding or from the average-pooled embedding within the segment.

\textbf{Results.} As shown in Fig.~\ref{fig:selective_decoding} (right), we sweep the average decoding frequency from 2.0 Hz down to 0.01 Hz (\textit{i.e.,} average intervals between consecutive decoding operations from 0.5s to 100s) by adjusting either the stride of uniform sampling or the number of clusters in adaptive selection. Across the entire range, adaptive selection consistently Pareto-dominates uniform sampling. In particular, selective decoding at 0.35 Hz (\textit{i.e.,} $\sim$2.85s interval) matches the performance of uniform decoding at 1 Hz, reducing decoding cost by $\sim$2.85$\times$. We further observe that average pooling provides consistent gains for both strategies, since it provides denoising and stabilization on embeddings prior feeding into the decoder.

\subsection{Evaluation of Y-Encoder}
\label{sec:exp.y-encoder}

\textbf{Evaluation Setup.} We evaluate whether the JEPA architecture improves the \texttt{Y-Encoder} by following the uni-modal text-only (TOT) evaluation setup.  
We use the hard-negative benchmarks SugarCrepe++ \citep{10.5555/3737916.3738487} and VISLA \citep{dumpala2024vislabenchmarkevaluatingembedding}.  
These datasets test sensitivity to semantic and lexical changes in image descriptions.  
Each dataset contains triplets: two semantically similar descriptions of the same image ($p1$ and $p2$), and one negative description ($n$) created by altering attributes, relations, or objects.  
We compare \texttt{Y-Encoders} from different models by computing the cosine similarity for all description pairs.  
We check that the similarity between positives $sim(p1,p2)$ is higher than both the similarity between each positive and the negative $sim(p1,n)$ and $sim(p2,n)$.  
We report accuracy (\%) across all samples.

\textbf{Results.} Table~\ref{tab:y-decoder} shows the performance of different models on text hard-negative benchmarks.
VL-JEPA$_\texttt{BASE}$ achieves a micro average accuracy of $63.9\%$ on SugarCrepe++ and $42.9\%$ on VISLA.
This is higher than the best other models: PE-Core scores $58.6\%$ on SugarCrepe++ and SigLIP2 scores $40.4\%$ on VISLA.
The finetuned VL-JEPA$_\texttt{SFT}$ model also achieves competitive results, with $58.4\%$ on SugarCrepe++ and $39.5\%$ on VISLA.
These results indicate that VL-JEPA$_\texttt{BASE}$ has a \texttt{Y-Encoder} that is more resilient to text hard-negatives.

\begin{table*}[t]
    \caption{\textbf{Ablation studies results}. The default setting adopted by VL-JEPA is marked in \colorbox{headerblue}{blue}. We calculate $\pm$delta within each group of ablations in comparison with the default setting.}
    \label{tab:ablations}
    \centering
    
    \begin{minipage}[t]{0.455\textwidth}
        \resizebox{\linewidth}{!}{%
        \begin{tabular}{r r l r l r l}
            \toprule
             & \multicolumn{2}{c}{\begin{tabular}{@{}c@{}}\textbf{Classification}\\(Accuracy)\end{tabular}} & \multicolumn{2}{c}{\begin{tabular}{@{}c@{}}\textbf{Retrieval}\\(Recall@1)\end{tabular}} & \multicolumn{2}{c}{\begin{tabular}{@{}c@{}}{\textbf{VQA}}\\(Accuracy)\end{tabular}} \\
            \midrule
            
            {\color{mygray} VL-JEPA$_{\texttt{SFT}}$} & {\color{mygray} 75.4} & & {\color{mygray} 63.8} & & {\color{mygray} 74.2} & \\
            \bottomrule \toprule
            
            \multicolumn{7}{l}{\textit{\color{mygray} (a) Effectiveness of pretraining stage on caption data}} \\
            \rowcolor{headerblue}
            w/ Pretraining & 49.0 & & 47.5 & & 46.1 & \\
            w/o Pretraining & 27.3 & \negval{-21.7} & 30.2 & \negval{-17.3} & 42.5 & \negval{-3.6} \\
            \bottomrule \toprule
            
            \multicolumn{7}{l}{\textit{\color{mygray} (b) Learning rate multiplier for Y-Encoder}} \\
            \rowcolor{headerblue}
            \textit{multiplier} = 0.05 & 27.3 & & 30.2 & & 42.5 & \\
            \textit{multiplier} = 1.00 & 23.7 & \negval{-3.6} & 28.8 & \negval{-1.4} & 40.7 & \negval{-1.8} \\
            \textit{multiplier} = 0.10 & 26.9 & \negval{-0.4} & 30.2 & \negval{-0.0} & 42.9 & \posval{+0.4} \\
            \textit{multiplier} = 0.01 & 25.6 & \negval{-1.7} & 27.7 & \negval{-2.5} & 41.0 & \negval{-1.5} \\
            \textit{multiplier} = 0.00 & 20.0 & \negval{-7.3} & 25.9 & \negval{-4.3} & 41.4 & \negval{-1.1} \\
            \bottomrule \toprule
            
            \multicolumn{7}{l}{\textit{\color{mygray} (c) Loss function (with no projection head on top frozen text encoder)}} \\
            \rowcolor{headerblue}
            InfoNCE & 23.3 & & 30.3 & & 44.3 & \\
            Cosine & 16.5 & \negval{-6.8} & 20.2 & \negval{-10.1} & 46.6 & \posval{+2.3} \\
            L1 & 14.8 & \negval{-8.5} & 15.5 & \negval{-14.8} & 41.9 & \negval{-2.4} \\
            L2 & 13.5 & \negval{-9.8} & 11.7 & \negval{-18.6} & 43.7 & \negval{-0.6} \\
            \bottomrule
        \end{tabular}%
        }
    \end{minipage}
    \hfill
    \begin{minipage}[t]{0.525\textwidth}
        \resizebox{\linewidth}{!}{%
        \begin{tabular}{r r l r l r l}
            \toprule
             & \multicolumn{2}{c}{\begin{tabular}{@{}c@{}}\textbf{Classification}\\(Accuracy)\end{tabular}} & \multicolumn{2}{c}{\begin{tabular}{@{}c@{}}\textbf{Retrieval}\\(Recall@1)\end{tabular}} & \multicolumn{2}{c}{\begin{tabular}{@{}c@{}}{\textbf{VQA}}\\(Accuracy)\end{tabular}} \\
            \midrule
            
            \multicolumn{7}{l}{\textit{\color{mygray} (d) Predictor architecture and initialization}} \\
            \rowcolor{headerblue}
            Layer 8-16 & 27.3 & & 30.2 & & 42.5 & \\
            Layer 0-2 & 24.3 & \negval{-3.0} & 27.8 & \negval{-2.4} & 40.1 & \negval{-2.4} \\
            Layer 0-4 & 25.1 & \negval{-2.2} & 28.9 & \negval{-1.3} & 43.6 & \posval{+1.1} \\
            Layer 0-8 & 27.2 & \negval{-0.1} & 29.3 & \negval{-0.9} & 43.4 & \posval{+0.9} \\
            Layer 0-16 & 27.4 & \posval{+0.1} & 31.0 & \posval{+0.8} & 45.5 & \posval{+3.0} \\
            w/o Bi-direction Attention & 26.7 & \negval{-0.6} & 31.2 & \posval{+1.0} & 40.6 & \negval{-1.9} \\
            w/o Llama-3 Initialization & 28.1 & \posval{+0.8} & 30.4 & \posval{+0.2} & 40.6 & \negval{-1.9} \\
            \bottomrule \toprule
            
            \multicolumn{7}{l}{\textit{\color{mygray} (e) Y-Encoder (trainable linear projection on top of frozen text encoder)}} \\
            \rowcolor{headerblue}
            EmbeddingGemma-300M & 19.5 & & 24.1 & & 42.5 & \\
            Qwen3-Embedding-0.6B & 24.5 & \posval{+5.0} & 24.5 & \posval{+0.4} & 41.5 & \negval{-1.0} \\
            Qwen3-Embedding-4B & 27.7 & \posval{+8.2} & 26.6 & \posval{+2.5} & 38.1 & \negval{-4.4} \\
            Qwen3-Embedding-8B & 29.6 & \posval{+10.1} & 29.5 & \posval{+5.4} & 41.9 & \negval{-0.6} \\
            PE$_{\text{core}}$-B (356M) & 29.4 & \posval{+9.9} & 34.5 & \posval{+10.4} & 35.9 & \negval{-6.6} \\
            PE$_{\text{core}}$-L (356M) & 29.0 & \posval{+9.5} & 34.2 & \posval{+10.1} & 42.9 & \posval{+0.4} \\
            PE$_{\text{core}}$-G (539M) & 33.9 & \posval{+14.4} & 32.0 & \posval{+7.9} & 41.8 & \negval{-0.7} \\
            \bottomrule
        \end{tabular}%
        }
    \end{minipage}
\end{table*}

\subsection{Ablation Study}
\label{sec:exp.ablations}

\textbf{Evaluation Setup.} We study different design choices for VL-JEPA. Here we train all ablation models on the SFT stage data for 10K steps with a batch size of 512 (5M samples seen) and constant learning rate. We report average classification top-1 accuracy of 8 datasets (Tab.~\ref{tab:cls_ret}), average text-to-video retrieval recall@1 of 8 datasets (Tab.~\ref{tab:cls_ret}), and {average VQA accuracy of 4 datasets (CLEVR, GQA, TallyQA simple and complex). We report the results in Tab.~\ref{tab:ablations}.}

\textbf{Results.} 
{\textbf{(a) Pretraining.} 
    Dropping the first query-free pretraining stage on image and video captions significantly hurt performance, especially on classification (-21.7) and retrieval (-17.3). 
\textbf{(b) LR Multiplier.} 
    The sweet point of learning rate multiplier to the Y-Encoder is around 0.05 to 0.10. Either faster or slower learning degrades the performance.
\textbf{(c) Loss Function.} 
    InfoNCE generally give superior performance compared to cosine, L1, and L2 losses, with the only exception being cosine loss outperform InfoNCE on VQA. However, only InfoNCE has the anti-collapse regularization and can be applied with unfrozen Y-Encoder.}
\textbf{(d) Predictor.} 
    In terms of predictor size, more layers yield better performance, especially on VQA performance. We also see that if using the original causal attention instead of updating to bi-direction attention hurt VQA performance (-1.9), since query tokens are appended after visual tokens, and visual tokens are no longer able to attend to query tokens. Finally, we also see that LLama-3 initialization is beneficial to VQA performance, although vision-language alignment (classification and retrieval) is a bit worse compared to randomly initialized Transformer layers.
{\textbf{(e) Y-Encoder.} 
    We tried different text encoder as the Y-Encoder, and confirmed that VL-JEPA works well with other embedding models than EmbeddingGemma-300M. Generally, larger encoder leads to better performance, with visually aligned text encoders (PE models) has significant advantage in classification and retrieval.}

%% file: content/related.tex
\section{Related Works}
\label{sec:related-works}

\textbf{JEPA Models.}  JEPA model learns by predicting the representation of a target input $Y$ from the representation of a context input $X$. Early instantiations include I-JEPA for image encoding \citep{assran2023self} and V-JEPA for video encoding \citep{bardes2023v}, which demonstrated the effectiveness of this objective over pixel reconstruction approach in their respective modality.
Recent JEPA work falls into two categories. One category of work emphasizes better unimodal representation learning \citep{assran2023self, bardes2023v, fei2023jepa} or cross-modal alignment \citep{lei2025m3, jose2025dinotxt}. 
The other direction targets world modeling, where pretrained encoders are frozen and action-conditioned predictors are trained for conditional prediction of state representations \citep{zhou2025dino,baldassarre2025back,assran2025v,terver2025drives}. This has shown good results but remains limited to narrow domains like mazes or robotic pick-and-place. 
Our proposed VL-JEPA is the first designed for general-purpose vision–language tasks. It performs conditional latent prediction over vision and text, and preserves efficiency while enabling flexible, multitask architecture.

\textbf{Vision Language Models.} Existing vision-language models largely fall into two families: (1) CLIP-style models with a non-predictive joint-embedding architecture (JEA) \citep{radford2021learning, zhai2023sigmoid, bolya2025perception, liu2024remoteclip, chen2023protoclip} encode images and texts independently into a common latent space, $X_V \!\mapsto\! S_V$ and $Y \!\mapsto\! S_Y$. By minimizing $\mathcal{L}_\texttt{CLIP} = D(S_V, S_Y)$ with a contrastive loss (\textit{e.g.,} InfoNCE), CLIP learns aligned \textit{representations} that support zero-shot classification and vision–language retrieval; (2) Generative VLMs \citep{liu2023visual, chen2022pali, dai2023instructblip, alayrac2022flamingo, chen2024visual, cho2025perceptionlm, beyer2024paligemma} connect a vision encoder \citep{radford2021learning,fini2025multimodal} with a language model (\textit{e.g.,} LLM). They are typically trained with $\mathcal{L}_\texttt{VLM} = D(\hat{Y}, Y)$, \textit{i.e.,} next token prediction with cross-entropy loss, and can learn to handle various vision-text-to-text generation tasks such as VQA.





\begin{table}[h!]
    \centering
    \caption{Task coverage comparison.}
    \label{tab:task_coverage}
    \resizebox{0.65\linewidth}{!}{%
    \begin{tabular}{cccc}
    \toprule
     & CLIP & VLM & VL-JEPA \\
    \midrule
    Generation       & \xmark & \cmark & \cmark \\
    Retrieval        & \cmark & \xmark & \cmark \\
    \bottomrule
    \end{tabular}}
\end{table}

Our proposed VL-JEPA integrates the architectural advantages and task coverage of both CLIPs and VLMs (Table~\ref{tab:task_coverage}). Since VL-JEPA learns in embedding space, it can leverage web-scale noisy image–text pairs \citep{jia2021scaling}, yielding strong open-domain features. On the other hand, VL-JEPA supports conditional generation tasks with a readout text decoder. Meanwhile, compared to generative VLMs that optimize directly in data space, VL-JEPA is more efficient at learning in the latent space. In addition, it is also more efficient for online inference, as it allows naturally selective decoding.

\textbf{Efficient Vision Language Models.} The growing size and training cost of VLMs has motivated efforts to improve efficiency. On the training side, strong performance can be achieved by updating only a subset of parameters, such as the vision–language connector \citep{tsimpoukelli2021multimodal,alayrac2022flamingo,vallaeys2024improved,shukor2023ep,koh2023groundingfromage,merullo2023linearlylimber,dai2023instructblip}. At inference, efficiency is pursued through pruning parameters or visual tokens \citep{cao2023pumer,shukor2024skipping,vasu2025fastvlm}. For real-time use cases, recent work explores small VLMs \citep{yao2024minicpm,marafioti2025smolvlm} and heuristics to reduce query frequency in asynchronous inference \citep{shukor2025smolvla}.

{
\textbf{Latent-space Language Modeling.}
Current state-of-the-art LLMs are trained to decode and reason in text space using autoregressive generation and chain-of-thought prompting \citep{wei2023chainofthoughtpromptingelicitsreasoning}.
Text-space LLMs have rapidly improved and now achieve strong results on a wide range of benchmarks.
However, the discrete nature of their reasoning trace may limit both speed and performance in the long term.
Several works have explored latent-space LLMs that process or reason in latent space, such as Large Concept Models \citep{lcmteam2024largeconceptmodelslanguage} and COCONUT \citep{hao2025traininglargelanguagemodels}.
These models focus on unimodal latent-space reasoning.
With VL-JEPA, our goal is to align vision and text representations in a shared multi-modal latent space.
This approach aims to enable better abstractions and improve both the performance and speed of vision-language models (VLMs).
We hope VL-JEPA will serve as a foundation for future work on multi-modal latent space reasoning, including visual chain-of-thought methods \citep{li2025imaginereasoningspacemultimodal}.
}

%% file: content/conclusion.tex
\section{Conclusion}
\label{sec:conclusion}

We have presented VL-JEPA, a new vision–language model built upon the joint embedding predictive architecture. By shifting supervision from discrete token space to continuous semantic embedding space, VL-JEPA simplifies the learning target, avoids redundant modeling of surface linguistic variability, and enables non-autoregressive prediction. Through controlled experiments, we show that VL-JEPA outperforms generative VLMs trained with cross-entropy loss under matched training data budget, while achieving superior training efficiency and significantly lower inference latency. Beyond generation tasks, the embedding-based design further allows VL-JEPA to handle open-vocabulary classification and cross-modal retrieval within a single unified architecture. Its ability to emit continuous semantic embeddings also makes it particularly well suited for real-time video applications, where selective decoding can improve both responsiveness and efficiency. 
In this work, we demonstrated the advantages of VL-JEPA over standard VLMs, particularly in computational efficiency, streaming applications, and video-language tasks. Our goal at this stage, is not to propose a universal alternative to VLMs, as this would require broader evaluation on tasks such as reasoning, tool use, and agentic behaviors where current token generative VLMs excel. Finally, although our results show clear benefits from scaling parameters and dataset size, we did not fully explore this direction, leaving it for future work.

\section*{Acknowledgments}

We would like to thank Loïc Barrault, Lucas Beyer, Quentin Garrido, João Maria Janeiro, Yifu Qiu, Koustuv Sinha, Basile Terver, and François Yvon for providing valuable feedback and support to this work. We thank Adrien Bardes for detailed V-JEPA 2 baseline performance on EK-100 action anticipation. We acknowledge Alejandro Castillejo Muñoz for his efforts on real-time demonstration and qualitative evaluation of VL-JEPA.

%% file: main.bbl
\begin{thebibliography}{85}
\providecommand{\natexlab}[1]{#1}
\providecommand{\url}[1]{\texttt{#1}}
\expandafter\ifx\csname urlstyle\endcsname\relax
  \providecommand{\doi}[1]{doi: #1}\else
  \providecommand{\doi}{doi: \begingroup \urlstyle{rm}\Url}\fi

\bibitem[Acharya et~al.(2019)Acharya, Kafle, and Kanan]{acharya2019tallyqa}
Manoj Acharya, Kushal Kafle, and Christopher Kanan.
\newblock Tallyqa: Answering complex counting questions.
\newblock In \emph{Proceedings of the AAAI conference on artificial intelligence}, volume~33, pages 8076--8084, 2019.

\bibitem[Alayrac et~al.(2022)Alayrac, Donahue, Luc, Miech, Barr, Hasson, Lenc, Mensch, Millican, Reynolds, et~al.]{alayrac2022flamingo}
Jean-Baptiste Alayrac, Jeff Donahue, Pauline Luc, Antoine Miech, Iain Barr, Yana Hasson, Karel Lenc, Arthur Mensch, Katherine Millican, Malcolm Reynolds, et~al.
\newblock Flamingo: a visual language model for few-shot learning.
\newblock \emph{Advances in neural information processing systems}, 35:\penalty0 23716--23736, 2022.

\bibitem[Anne~Hendricks et~al.(2017)Anne~Hendricks, Wang, Shechtman, Sivic, Darrell, and Russell]{anne2017localizing}
Lisa Anne~Hendricks, Oliver Wang, Eli Shechtman, Josef Sivic, Trevor Darrell, and Bryan Russell.
\newblock Localizing moments in video with natural language.
\newblock In \emph{Proceedings of the IEEE international conference on computer vision}, pages 5803--5812, 2017.

\bibitem[Assran et~al.(2023)Assran, Duval, Misra, Bojanowski, Vincent, Rabbat, LeCun, and Ballas]{assran2023self}
Mahmoud Assran, Quentin Duval, Ishan Misra, Piotr Bojanowski, Pascal Vincent, Michael Rabbat, Yann LeCun, and Nicolas Ballas.
\newblock Self-supervised learning from images with a joint-embedding predictive architecture.
\newblock In \emph{Proceedings of the IEEE/CVF Conference on Computer Vision and Pattern Recognition}, pages 15619--15629, 2023.

\bibitem[Assran et~al.(2025)Assran, Bardes, Fan, Garrido, Howes, Muckley, Rizvi, Roberts, Sinha, Zholus, et~al.]{assran2025v}
Mido Assran, Adrien Bardes, David Fan, Quentin Garrido, Russell Howes, Matthew Muckley, Ammar Rizvi, Claire Roberts, Koustuv Sinha, Artem Zholus, et~al.
\newblock V-jepa 2: Self-supervised video models enable understanding, prediction and planning.
\newblock \emph{arXiv preprint arXiv:2506.09985}, 2025.

\bibitem[Bai et~al.(2023)Bai, Bai, Chu, Cui, Dang, Deng, Fan, Ge, Han, Huang, et~al.]{bai2023qwen}
Jinze Bai, Shuai Bai, Yunfei Chu, Zeyu Cui, Kai Dang, Xiaodong Deng, Yang Fan, Wenbin Ge, Yu~Han, Fei Huang, et~al.
\newblock Qwen technical report.
\newblock \emph{arXiv preprint arXiv:2309.16609}, 2023.

\bibitem[Baldassarre et~al.(2025)Baldassarre, Szafraniec, Terver, Khalidov, Massa, LeCun, Labatut, Seitzer, and Bojanowski]{baldassarre2025back}
Federico Baldassarre, Marc Szafraniec, Basile Terver, Vasil Khalidov, Francisco Massa, Yann LeCun, Patrick Labatut, Maximilian Seitzer, and Piotr Bojanowski.
\newblock Back to the features: Dino as a foundation for video world models.
\newblock \emph{arXiv preprint arXiv:2507.19468}, 2025.

\bibitem[Balestriero and LeCun(2025)]{balestriero2025lejepa}
Randall Balestriero and Yann LeCun.
\newblock Lejepa: Provable and scalable self-supervised learning without the heuristics.
\newblock \emph{arXiv preprint arXiv:2511.08544}, 2025.

\bibitem[Bardes et~al.(2021)Bardes, Ponce, and LeCun]{bardes2021vicreg}
Adrien Bardes, Jean Ponce, and Yann LeCun.
\newblock Vicreg: Variance-invariance-covariance regularization for self-supervised learning.
\newblock \emph{arXiv preprint arXiv:2105.04906}, 2021.

\bibitem[Bardes et~al.(2023)Bardes, Garrido, Ponce, Chen, Rabbat, LeCun, Assran, and Ballas]{bardes2023v}
Adrien Bardes, Quentin Garrido, Jean Ponce, Xinlei Chen, Michael Rabbat, Yann LeCun, Mido Assran, and Nicolas Ballas.
\newblock V-jepa: Latent video prediction for visual representation learning.
\newblock 2023.

\bibitem[Barrault et~al.(2024)Barrault, Duquenne, Elbayad, Kozhevnikov, Alastruey, Andrews, Coria, Couairon, Costa-juss{\`a}, Dale, et~al.]{lcmteam2024largeconceptmodelslanguage}
Lo{\"\i}c Barrault, Paul-Ambroise Duquenne, Maha Elbayad, Artyom Kozhevnikov, Belen Alastruey, Pierre Andrews, Mariano Coria, Guillaume Couairon, Marta~R Costa-juss{\`a}, David Dale, et~al.
\newblock Large concept models: Language modeling in a sentence representation space.
\newblock \emph{arXiv preprint arXiv:2412.08821}, 2024.

\bibitem[Beyer et~al.(2024)Beyer, Steiner, Pinto, Kolesnikov, Wang, Salz, Neumann, Alabdulmohsin, Tschannen, Bugliarello, et~al.]{beyer2024paligemma}
Lucas Beyer, Andreas Steiner, Andr{\'e}~Susano Pinto, Alexander Kolesnikov, Xiao Wang, Daniel Salz, Maxim Neumann, Ibrahim Alabdulmohsin, Michael Tschannen, Emanuele Bugliarello, et~al.
\newblock Paligemma: A versatile 3b vlm for transfer.
\newblock \emph{arXiv preprint arXiv:2407.07726}, 2024.

\bibitem[Black et~al.(2025)Black, Galliker, and Levine]{black2025real}
Kevin Black, Manuel~Y Galliker, and Sergey Levine.
\newblock Real-time execution of action chunking flow policies.
\newblock \emph{arXiv preprint arXiv:2506.07339}, 2025.

\bibitem[Bolya et~al.(2025)Bolya, Huang, Sun, Cho, Madotto, Wei, Ma, Zhi, Rajasegaran, Rasheed, et~al.]{bolya2025perception}
Daniel Bolya, Po-Yao Huang, Peize Sun, Jang~Hyun Cho, Andrea Madotto, Chen Wei, Tengyu Ma, Jiale Zhi, Jathushan Rajasegaran, Hanoona Rasheed, et~al.
\newblock Perception encoder: The best visual embeddings are not at the output of the network.
\newblock \emph{arXiv preprint arXiv:2504.13181}, 2025.

\bibitem[Bordes et~al.(2024)Bordes, Pang, Ajay, Li, Bardes, Petryk, Ma{\~n}as, Lin, Mahmoud, Jayaraman, et~al.]{bordes2024introduction}
Florian Bordes, Richard~Yuanzhe Pang, Anurag Ajay, Alexander~C Li, Adrien Bardes, Suzanne Petryk, Oscar Ma{\~n}as, Zhiqiu Lin, Anas Mahmoud, Bargav Jayaraman, et~al.
\newblock An introduction to vision-language modeling.
\newblock \emph{arXiv preprint arXiv:2405.17247}, 2024.

\bibitem[Caba~Heilbron et~al.(2015)Caba~Heilbron, Escorcia, Ghanem, and Carlos~Niebles]{caba2015activitynet}
Fabian Caba~Heilbron, Victor Escorcia, Bernard Ghanem, and Juan Carlos~Niebles.
\newblock Activitynet: A large-scale video benchmark for human activity understanding.
\newblock In \emph{Proceedings of the ieee conference on computer vision and pattern recognition}, pages 961--970, 2015.

\bibitem[Cao et~al.(2023)Cao, Paranjape, and Hajishirzi]{cao2023pumer}
Qingqing Cao, Bhargavi Paranjape, and Hannaneh Hajishirzi.
\newblock Pumer: Pruning and merging tokens for efficient vision language models.
\newblock \emph{arXiv preprint arXiv:2305.17530}, 2023.

\bibitem[Chai et~al.(2024)Chai, Song, Du, Meng, Madhavan, Bar-Tal, Hwang, Xie, and Manning]{chai2024auroracap}
Wenhao Chai, Enxin Song, Yilun Du, Chenlin Meng, Vashisht Madhavan, Omer Bar-Tal, Jenq-Neng Hwang, Saining Xie, and Christopher~D Manning.
\newblock Auroracap: Efficient, performant video detailed captioning and a new benchmark.
\newblock \emph{arXiv preprint arXiv:2410.03051}, 2024.

\bibitem[Chatterjee et~al.(2025)Chatterjee, Remelli, Song, Tekin, Mittal, Bhatnagar, Camg{\~A}{\c{k}}z, Hampali, Sauser, Ma, et~al.]{chatterjee2025memory}
Dibyadip Chatterjee, Edoardo Remelli, Yale Song, Bugra Tekin, Abhay Mittal, Bharat Bhatnagar, Necati~Cihan Camg{\~A}{\c{k}}z, Shreyas Hampali, Eric Sauser, Shugao Ma, et~al.
\newblock Memory-efficient streaming videollms for real-time procedural video understanding.
\newblock \emph{arXiv preprint arXiv:2504.13915}, 2025.

\bibitem[Chen and Dolan(2011)]{chen2011collecting}
David Chen and William~B Dolan.
\newblock Collecting highly parallel data for paraphrase evaluation.
\newblock In \emph{Proceedings of the 49th annual meeting of the association for computational linguistics: human language technologies}, pages 190--200, 2011.

\bibitem[Chen et~al.(2023)Chen, Wu, Liu, Yang, Zheng, Tan, and Zhou]{chen2023protoclip}
Delong Chen, Zhao Wu, Fan Liu, Zaiquan Yang, Shaoqiu Zheng, Ying Tan, and Erjin Zhou.
\newblock Protoclip: Prototypical contrastive language image pretraining.
\newblock \emph{IEEE Transactions on Neural Networks and Learning Systems}, 2023.

\bibitem[Chen et~al.(2024{\natexlab{a}})Chen, Cahyawijaya, Liu, Wang, and Fung]{chen2024subobject}
Delong Chen, Samuel Cahyawijaya, Jianfeng Liu, Baoyuan Wang, and Pascale Fung.
\newblock Subobject-level image tokenization.
\newblock \emph{arXiv preprint arXiv:2402.14327}, 2024{\natexlab{a}}.

\bibitem[Chen et~al.(2024{\natexlab{b}})Chen, Liu, Dai, and Wang]{chen2024visual}
Delong Chen, Jianfeng Liu, Wenliang Dai, and Baoyuan Wang.
\newblock Visual instruction tuning with polite flamingo.
\newblock In \emph{Proceedings of the aaai conference on artificial intelligence}, volume~38, pages 17745--17753, 2024{\natexlab{b}}.

\bibitem[Chen et~al.(2025{\natexlab{a}})Chen, Chung, Bang, Ji, and Fung]{chen2025worldprediction}
Delong Chen, Willy Chung, Yejin Bang, Ziwei Ji, and Pascale Fung.
\newblock Worldprediction: A benchmark for high-level world modeling and long-horizon procedural planning.
\newblock \emph{arXiv preprint arXiv:2506.04363}, 2025{\natexlab{a}}.

\bibitem[Chen et~al.(2025{\natexlab{b}})Chen, Moutakanni, Chung, Bang, Ji, Bolourchi, and Fung]{chen2025planning}
Delong Chen, Theo Moutakanni, Willy Chung, Yejin Bang, Ziwei Ji, Allen Bolourchi, and Pascale Fung.
\newblock Planning with reasoning using vision language world model.
\newblock \emph{arXiv preprint arXiv:2509.02722}, 2025{\natexlab{b}}.

\bibitem[Chen et~al.(2026)Chen, Kasarla, Bang, Shukor, Chung, Yu, Bolourchi, Moutakanni, and Fung]{chen2026action100m}
Delong Chen, Tejaswi Kasarla, Yejin Bang, Mustafa Shukor, Willy Chung, Jade Yu, Allen Bolourchi, Theo Moutakanni, and Pascale Fung.
\newblock Action100m: A large-scale video action dataset.
\newblock \emph{arXiv preprint arXiv:2601.10592}, 2026.

\bibitem[Chen et~al.(2024{\natexlab{c}})Chen, Lv, Wu, Lin, Song, Gao, Liu, Gao, Mao, and Shou]{chen2024videollm}
Joya Chen, Zhaoyang Lv, Shiwei Wu, Kevin~Qinghong Lin, Chenan Song, Difei Gao, Jia-Wei Liu, Ziteng Gao, Dongxing Mao, and Mike~Zheng Shou.
\newblock Videollm-online: Online video large language model for streaming video.
\newblock In \emph{Proceedings of the IEEE/CVF Conference on Computer Vision and Pattern Recognition}, pages 18407--18418, 2024{\natexlab{c}}.

\bibitem[Chen et~al.(2022)Chen, Wang, Changpinyo, Piergiovanni, Padlewski, Salz, Goodman, Grycner, Mustafa, Beyer, et~al.]{chen2022pali}
Xi~Chen, Xiao Wang, Soravit Changpinyo, Anthony~J Piergiovanni, Piotr Padlewski, Daniel Salz, Sebastian Goodman, Adam Grycner, Basil Mustafa, Lucas Beyer, et~al.
\newblock Pali: A jointly-scaled multilingual language-image model.
\newblock \emph{arXiv preprint arXiv:2209.06794}, 2022.

\bibitem[Chen et~al.(2024{\natexlab{d}})Chen, Wu, Wang, Su, Chen, Xing, Zhong, Zhang, Zhu, Lu, et~al.]{chen2024internvl}
Zhe Chen, Jiannan Wu, Wenhai Wang, Weijie Su, Guo Chen, Sen Xing, Muyan Zhong, Qinglong Zhang, Xizhou Zhu, Lewei Lu, et~al.
\newblock Internvl: Scaling up vision foundation models and aligning for generic visual-linguistic tasks.
\newblock In \emph{Proceedings of the IEEE/CVF conference on computer vision and pattern recognition}, pages 24185--24198, 2024{\natexlab{d}}.

\bibitem[Cho et~al.(2025)Cho, Madotto, Mavroudi, Afouras, Nagarajan, Maaz, Song, Ma, Hu, Jain, et~al.]{cho2025perceptionlm}
Jang~Hyun Cho, Andrea Madotto, Effrosyni Mavroudi, Triantafyllos Afouras, Tushar Nagarajan, Muhammad Maaz, Yale Song, Tengyu Ma, Shuming Hu, Suyog Jain, et~al.
\newblock Perceptionlm: Open-access data and models for detailed visual understanding.
\newblock \emph{arXiv preprint arXiv:2504.13180}, 2025.

\bibitem[Dai et~al.(2023)Dai, Li, Li, Tiong, Zhao, Wang, Li, Fung, and Hoi]{dai2023instructblip}
Wenliang Dai, Junnan Li, Dongxu Li, Anthony Tiong, Junqi Zhao, Weisheng Wang, Boyang Li, Pascale~N Fung, and Steven Hoi.
\newblock Instructblip: Towards general-purpose vision-language models with instruction tuning.
\newblock \emph{Advances in neural information processing systems}, 36:\penalty0 49250--49267, 2023.

\bibitem[Damen et~al.(2022)Damen, Doughty, Farinella, Furnari, Kazakos, Ma, Moltisanti, Munro, Perrett, Price, et~al.]{damen2022rescaling}
Dima Damen, Hazel Doughty, Giovanni~Maria Farinella, Antonino Furnari, Evangelos Kazakos, Jian Ma, Davide Moltisanti, Jonathan Munro, Toby Perrett, Will Price, et~al.
\newblock Rescaling egocentric vision: Collection, pipeline and challenges for epic-kitchens-100.
\newblock \emph{International Journal of Computer Vision}, 130\penalty0 (1):\penalty0 33--55, 2022.

\bibitem[Di et~al.(2025)Di, Yu, Zhang, Li, Zhong, Cheng, Li, He, Shu, and Jiang]{di2025streaming}
Shangzhe Di, Zhelun Yu, Guanghao Zhang, Haoyuan Li, Tao Zhong, Hao Cheng, Bolin Li, Wanggui He, Fangxun Shu, and Hao Jiang.
\newblock Streaming video question-answering with in-context video kv-cache retrieval.
\newblock \emph{arXiv preprint arXiv:2503.00540}, 2025.

\bibitem[Dumpala et~al.(2024{\natexlab{a}})Dumpala, Jaiswal, Sastry, Milios, Oore, and Sajjad]{10.5555/3737916.3738487}
Sri~Harsha Dumpala, Aman Jaiswal, Chandramouli Sastry, Evangelos Milios, Sageev Oore, and Hassan Sajjad.
\newblock Sugarcrepe++ dataset: vision-language model sensitivity to semantic and lexical alterations.
\newblock In \emph{Proceedings of the 38th International Conference on Neural Information Processing Systems}, NIPS '24, Red Hook, NY, USA, 2024{\natexlab{a}}. Curran Associates Inc.
\newblock ISBN 9798331314385.

\bibitem[Dumpala et~al.(2024{\natexlab{b}})Dumpala, Jaiswal, Sastry, Milios, Oore, and Sajjad]{dumpala2024vislabenchmarkevaluatingembedding}
Sri~Harsha Dumpala, Aman Jaiswal, Chandramouli Sastry, Evangelos Milios, Sageev Oore, and Hassan Sajjad.
\newblock Visla benchmark: Evaluating embedding sensitivity to semantic and lexical alterations.
\newblock \emph{arXiv preprint arXiv:2404.16365}, 2024{\natexlab{b}}.

\bibitem[Duquenne et~al.(2023)Duquenne, Schwenk, and Sagot]{duquenne2023sonar}
Paul-Ambroise Duquenne, Holger Schwenk, and Beno{\^\i}t Sagot.
\newblock Sonar: sentence-level multimodal and language-agnostic representations.
\newblock \emph{arXiv preprint arXiv:2308.11466}, 2023.

\bibitem[Fei et~al.(2023)Fei, Fan, and Huang]{fei2023jepa}
Zhengcong Fei, Mingyuan Fan, and Junshi Huang.
\newblock A-jepa: Joint-embedding predictive architecture can listen.
\newblock \emph{arXiv preprint arXiv:2311.15830}, 2023.

\bibitem[Fini et~al.(2025)Fini, Shukor, Li, Dufter, Klein, Haldimann, Aitharaju, da~Costa, B{\'e}thune, Gan, et~al.]{fini2025multimodal}
Enrico Fini, Mustafa Shukor, Xiujun Li, Philipp Dufter, Michal Klein, David Haldimann, Sai Aitharaju, Victor G~Turrisi da~Costa, Louis B{\'e}thune, Zhe Gan, et~al.
\newblock Multimodal autoregressive pre-training of large vision encoders.
\newblock In \emph{Proceedings of the Computer Vision and Pattern Recognition Conference}, pages 9641--9654, 2025.

\bibitem[Fung et~al.(2025)Fung, Bachrach, Celikyilmaz, Chaudhuri, Chen, Chung, Dupoux, Gong, J{\'e}gou, Lazaric, et~al.]{fung2025embodied}
Pascale Fung, Yoram Bachrach, Asli Celikyilmaz, Kamalika Chaudhuri, Delong Chen, Willy Chung, Emmanuel Dupoux, Hongyu Gong, Herv{\'e} J{\'e}gou, Alessandro Lazaric, et~al.
\newblock Embodied ai agents: Modeling the world.
\newblock \emph{arXiv preprint arXiv:2506.22355}, 2025.

\bibitem[Gadre et~al.(2023)Gadre, Ilharco, Fang, Hayase, Smyrnis, Nguyen, Marten, Wortsman, Ghosh, Zhang, et~al.]{gadre2023datacomp}
Samir~Yitzhak Gadre, Gabriel Ilharco, Alex Fang, Jonathan Hayase, Georgios Smyrnis, Thao Nguyen, Ryan Marten, Mitchell Wortsman, Dhruba Ghosh, Jieyu Zhang, et~al.
\newblock Datacomp: In search of the next generation of multimodal datasets.
\newblock \emph{Advances in Neural Information Processing Systems}, 36:\penalty0 27092--27112, 2023.

\bibitem[Goyal et~al.(2017)Goyal, Ebrahimi~Kahou, Michalski, Materzynska, Westphal, Kim, Haenel, Fruend, Yianilos, Mueller-Freitag, et~al.]{goyal2017something}
Raghav Goyal, Samira Ebrahimi~Kahou, Vincent Michalski, Joanna Materzynska, Susanne Westphal, Heuna Kim, Valentin Haenel, Ingo Fruend, Peter Yianilos, Moritz Mueller-Freitag, et~al.
\newblock The" something something" video database for learning and evaluating visual common sense.
\newblock In \emph{Proceedings of the IEEE international conference on computer vision}, pages 5842--5850, 2017.

\bibitem[Grauman et~al.(2024)Grauman, Westbury, Torresani, Kitani, Malik, Afouras, Ashutosh, Baiyya, Bansal, Boote, et~al.]{grauman2024ego}
Kristen Grauman, Andrew Westbury, Lorenzo Torresani, Kris Kitani, Jitendra Malik, Triantafyllos Afouras, Kumar Ashutosh, Vijay Baiyya, Siddhant Bansal, Bikram Boote, et~al.
\newblock Ego-exo4d: Understanding skilled human activity from first-and third-person perspectives.
\newblock In \emph{Proceedings of the IEEE/CVF Conference on Computer Vision and Pattern Recognition}, pages 19383--19400, 2024.

\bibitem[Hao et~al.(2024)Hao, Sukhbaatar, Su, Li, Hu, Weston, and Tian]{hao2025traininglargelanguagemodels}
Shibo Hao, Sainbayar Sukhbaatar, DiJia Su, Xian Li, Zhiting Hu, Jason Weston, and Yuandong Tian.
\newblock Training large language models to reason in a continuous latent space.
\newblock \emph{arXiv preprint arXiv:2412.06769}, 2024.

\bibitem[Hudson and Manning(2019)]{hudson2019gqa}
Drew~A Hudson and Christopher~D Manning.
\newblock Gqa: A new dataset for real-world visual reasoning and compositional question answering.
\newblock In \emph{Proceedings of the IEEE/CVF conference on computer vision and pattern recognition}, pages 6700--6709, 2019.

\bibitem[Jia et~al.(2021)Jia, Yang, Xia, Chen, Parekh, Pham, Le, Sung, Li, and Duerig]{jia2021scaling}
Chao Jia, Yinfei Yang, Ye~Xia, Yi-Ting Chen, Zarana Parekh, Hieu Pham, Quoc Le, Yun-Hsuan Sung, Zhen Li, and Tom Duerig.
\newblock Scaling up visual and vision-language representation learning with noisy text supervision.
\newblock In \emph{International conference on machine learning}, pages 4904--4916. PMLR, 2021.

\bibitem[Jose et~al.(2025)Jose, Moutakanni, Kang, Baldassarre, Darcet, Xu, Li, Szafraniec, Ramamonjisoa, Oquab, et~al.]{jose2025dinotxt}
Cijo Jose, Th{\'e}o Moutakanni, Dahyun Kang, Federico Baldassarre, Timoth{\'e}e Darcet, Hu~Xu, Daniel Li, Marc Szafraniec, Micha{\"e}l Ramamonjisoa, Maxime Oquab, et~al.
\newblock Dinov2 meets text: A unified framework for image-and pixel-level vision-language alignment.
\newblock In \emph{Proceedings of the Computer Vision and Pattern Recognition Conference}, pages 24905--24916, 2025.

\bibitem[Kay et~al.(2017)Kay, Carreira, Simonyan, Zhang, Hillier, Vijayanarasimhan, Viola, Green, Back, Natsev, et~al.]{kay2017kinetics}
Will Kay, Joao Carreira, Karen Simonyan, Brian Zhang, Chloe Hillier, Sudheendra Vijayanarasimhan, Fabio Viola, Tim Green, Trevor Back, Paul Natsev, et~al.
\newblock The kinetics human action video dataset.
\newblock \emph{arXiv preprint arXiv:1705.06950}, 2017.

\bibitem[Koh et~al.(2023)Koh, Salakhutdinov, and Fried]{koh2023groundingfromage}
Jing~Yu Koh, Ruslan Salakhutdinov, and Daniel Fried.
\newblock Grounding language models to images for multimodal inputs and outputs.
\newblock In \emph{International Conference on Machine Learning}, pages 17283--17300. PMLR, 2023.

\bibitem[LeCun(2022)]{lecun2022path}
Yann LeCun.
\newblock A path towards autonomous machine intelligence.
\newblock \emph{Open Review}, 62\penalty0 (1):\penalty0 1--62, 2022.

\bibitem[Lei et~al.(2025)Lei, Cheng, Qin, Wang, Huang, Gu, Wu, and Ji]{lei2025m3}
Hongyang Lei, Xiaolong Cheng, Qi~Qin, Dan Wang, Huazhen Huang, Qingqing Gu, Yetao Wu, and Luo Ji.
\newblock M3-jepa: Multimodal alignment via multi-gate moe based on the joint-embedding predictive architecture.
\newblock In \emph{Forty-second International Conference on Machine Learning}, 2025.

\bibitem[Li et~al.(2025{\natexlab{a}})Li, Wu, Zhang, Xia, Mao, Dong, Vuli{\'c}, and Wei]{li2025imaginereasoningspacemultimodal}
Chengzu Li, Wenshan Wu, Huanyu Zhang, Yan Xia, Shaoguang Mao, Li~Dong, Ivan Vuli{\'c}, and Furu Wei.
\newblock Imagine while reasoning in space: Multimodal visualization-of-thought.
\newblock \emph{arXiv preprint arXiv:2501.07542}, 2025{\natexlab{a}}.

\bibitem[Li et~al.(2023{\natexlab{a}})Li, Li, Savarese, and Hoi]{li2023blip}
Junnan Li, Dongxu Li, Silvio Savarese, and Steven Hoi.
\newblock Blip-2: Bootstrapping language-image pre-training with frozen image encoders and large language models.
\newblock In \emph{International conference on machine learning}, pages 19730--19742. PMLR, 2023{\natexlab{a}}.

\bibitem[Li et~al.(2023{\natexlab{b}})Li, Du, Zhou, Wang, Zhao, and Wen]{li2023evaluating}
Yifan Li, Yifan Du, Kun Zhou, Jinpeng Wang, Wayne~Xin Zhao, and Ji-Rong Wen.
\newblock Evaluating object hallucination in large vision-language models.
\newblock \emph{arXiv preprint arXiv:2305.10355}, 2023{\natexlab{b}}.

\bibitem[Li et~al.(2025{\natexlab{b}})Li, Zhou, Zhao, Fang, and Wen]{li2025analyzing}
Yifan Li, Kun Zhou, Wayne~Xin Zhao, Lei Fang, and Ji-Rong Wen.
\newblock Analyzing and mitigating object hallucination: A training bias perspective.
\newblock \emph{arXiv preprint arXiv:2508.04567}, 2025{\natexlab{b}}.

\bibitem[Lin et~al.(2024)Lin, Ye, Zhu, Cui, Ning, Jin, and Yuan]{lin2024video}
Bin Lin, Yang Ye, Bin Zhu, Jiaxi Cui, Munan Ning, Peng Jin, and Li~Yuan.
\newblock Video-llava: Learning united visual representation by alignment before projection.
\newblock In \emph{Proceedings of the 2024 Conference on Empirical Methods in Natural Language Processing}, pages 5971--5984, 2024.

\bibitem[Liu et~al.(2024)Liu, Chen, Guan, Zhou, Zhu, Ye, Fu, and Zhou]{liu2024remoteclip}
Fan Liu, Delong Chen, Zhangqingyun Guan, Xiaocong Zhou, Jiale Zhu, Qiaolin Ye, Liyong Fu, and Jun Zhou.
\newblock Remoteclip: A vision language foundation model for remote sensing.
\newblock \emph{IEEE Transactions on Geoscience and Remote Sensing}, 62:\penalty0 1--16, 2024.

\bibitem[Liu et~al.(2023)Liu, Li, Wu, and Lee]{liu2023visual}
Haotian Liu, Chunyuan Li, Qingyang Wu, and Yong~Jae Lee.
\newblock Visual instruction tuning.
\newblock \emph{Advances in neural information processing systems}, 36:\penalty0 34892--34916, 2023.

\bibitem[Marafioti et~al.(2025)Marafioti, Zohar, Farr{\'e}, Noyan, Bakouch, Cuenca, Zakka, Allal, Lozhkov, Tazi, et~al.]{marafioti2025smolvlm}
Andr{\'e}s Marafioti, Orr Zohar, Miquel Farr{\'e}, Merve Noyan, Elie Bakouch, Pedro Cuenca, Cyril Zakka, Loubna~Ben Allal, Anton Lozhkov, Nouamane Tazi, et~al.
\newblock Smolvlm: Redefining small and efficient multimodal models.
\newblock \emph{arXiv preprint arXiv:2504.05299}, 2025.

\bibitem[Merullo et~al.(2022)Merullo, Castricato, Eickhoff, and Pavlick]{merullo2023linearlylimber}
Jack Merullo, Louis Castricato, Carsten Eickhoff, and Ellie Pavlick.
\newblock Linearly mapping from image to text space.
\newblock \emph{arXiv preprint arXiv:2209.15162}, 2022.

\bibitem[Murtagh and Contreras(2012)]{murtagh2012algorithms}
Fionn Murtagh and Pedro Contreras.
\newblock Algorithms for hierarchical clustering: an overview.
\newblock \emph{Wiley interdisciplinary reviews: data mining and knowledge discovery}, 2\penalty0 (1):\penalty0 86--97, 2012.

\bibitem[Qian et~al.(2024)Qian, Dong, Zhang, Zang, Ding, Lin, and Wang]{qian2024streaming}
Rui Qian, Xiaoyi Dong, Pan Zhang, Yuhang Zang, Shuangrui Ding, Dahua Lin, and Jiaqi Wang.
\newblock Streaming long video understanding with large language models.
\newblock \emph{Advances in Neural Information Processing Systems}, 37:\penalty0 119336--119360, 2024.

\bibitem[Radford et~al.(2021)Radford, Kim, Hallacy, Ramesh, Goh, Agarwal, Sastry, Askell, Mishkin, Clark, et~al.]{radford2021learning}
Alec Radford, Jong~Wook Kim, Chris Hallacy, Aditya Ramesh, Gabriel Goh, Sandhini Agarwal, Girish Sastry, Amanda Askell, Pamela Mishkin, Jack Clark, et~al.
\newblock Learning transferable visual models from natural language supervision.
\newblock In \emph{International conference on machine learning}, pages 8748--8763. PmLR, 2021.

\bibitem[Shukor and Cord(2024)]{shukor2024skipping}
Mustafa Shukor and Matthieu Cord.
\newblock Skipping computations in multimodal llms.
\newblock \emph{arXiv preprint arXiv:2410.09454}, 2024.

\bibitem[Shukor et~al.(2023)Shukor, Dancette, and Cord]{shukor2023ep}
Mustafa Shukor, Corentin Dancette, and Matthieu Cord.
\newblock ep-alm: Efficient perceptual augmentation of language models.
\newblock In \emph{Proceedings of the IEEE/CVF International Conference on Computer Vision}, pages 22056--22069, 2023.

\bibitem[Shukor et~al.(2025)Shukor, Aubakirova, Capuano, Kooijmans, Palma, Zouitine, Aractingi, Pascal, Russi, Marafioti, et~al.]{shukor2025smolvla}
Mustafa Shukor, Dana Aubakirova, Francesco Capuano, Pepijn Kooijmans, Steven Palma, Adil Zouitine, Michel Aractingi, Caroline Pascal, Martino Russi, Andres Marafioti, et~al.
\newblock Smolvla: A vision-language-action model for affordable and efficient robotics.
\newblock \emph{arXiv preprint arXiv:2506.01844}, 2025.

\bibitem[Song et~al.(2025)Song, Chen, Ding, Zhao, Zhao, Zhong, Ge, Ma, and Li]{song2025accelerating}
Wenxuan Song, Jiayi Chen, Pengxiang Ding, Han Zhao, Wei Zhao, Zhide Zhong, Zongyuan Ge, Jun Ma, and Haoang Li.
\newblock Accelerating vision-language-action model integrated with action chunking via parallel decoding.
\newblock \emph{arXiv preprint arXiv:2503.02310}, 2025.

\bibitem[Tang et~al.(2019)Tang, Ding, Rao, Zheng, Zhang, Zhao, Lu, and Zhou]{tang2019coin}
Yansong Tang, Dajun Ding, Yongming Rao, Yu~Zheng, Danyang Zhang, Lili Zhao, Jiwen Lu, and Jie Zhou.
\newblock Coin: A large-scale dataset for comprehensive instructional video analysis.
\newblock In \emph{Proceedings of the IEEE/CVF Conference on Computer Vision and Pattern Recognition}, pages 1207--1216, 2019.

\bibitem[Terver et~al.(2025)Terver, Yang, Ponce, Bardes, and LeCun]{terver2025drives}
Basile Terver, Tsung-Yen Yang, Jean Ponce, Adrien Bardes, and Yann LeCun.
\newblock What drives success in physical planning with joint-embedding predictive world models?
\newblock \emph{arXiv preprint arXiv:2512.24497}, 2025.

\bibitem[Thomee et~al.(2016)Thomee, Shamma, Friedland, Elizalde, Ni, Poland, Borth, and Li]{thomee2016yfcc100m}
Bart Thomee, David~A Shamma, Gerald Friedland, Benjamin Elizalde, Karl Ni, Douglas Poland, Damian Borth, and Li-Jia Li.
\newblock Yfcc100m: The new data in multimedia research.
\newblock \emph{Communications of the ACM}, 59\penalty0 (2):\penalty0 64--73, 2016.

\bibitem[Tschannen et~al.(2025)Tschannen, Gritsenko, Wang, Naeem, Alabdulmohsin, Parthasarathy, Evans, Beyer, Xia, Mustafa, et~al.]{tschannen2025siglip}
Michael Tschannen, Alexey Gritsenko, Xiao Wang, Muhammad~Ferjad Naeem, Ibrahim Alabdulmohsin, Nikhil Parthasarathy, Talfan Evans, Lucas Beyer, Ye~Xia, Basil Mustafa, et~al.
\newblock Siglip 2: Multilingual vision-language encoders with improved semantic understanding, localization, and dense features.
\newblock \emph{arXiv preprint arXiv:2502.14786}, 2025.

\bibitem[Tsimpoukelli et~al.(2021)Tsimpoukelli, Menick, Cabi, Eslami, Vinyals, and Hill]{tsimpoukelli2021multimodal}
Maria Tsimpoukelli, Jacob~L Menick, Serkan Cabi, SM~Eslami, Oriol Vinyals, and Felix Hill.
\newblock Multimodal few-shot learning with frozen language models.
\newblock \emph{Advances in Neural Information Processing Systems}, 34:\penalty0 200--212, 2021.

\bibitem[Vallaeys et~al.(2024)Vallaeys, Shukor, Cord, and Verbeek]{vallaeys2024improved}
Th{\'e}ophane Vallaeys, Mustafa Shukor, Matthieu Cord, and Jakob Verbeek.
\newblock Improved baselines for data-efficient perceptual augmentation of llms.
\newblock In \emph{European Conference on Computer Vision}, pages 369--387. Springer, 2024.

\bibitem[Vasu et~al.(2025)Vasu, Faghri, Li, Koc, True, Antony, Santhanam, Gabriel, Grasch, Tuzel, et~al.]{vasu2025fastvlm}
Pavan Kumar~Anasosalu Vasu, Fartash Faghri, Chun-Liang Li, Cem Koc, Nate True, Albert Antony, Gokula Santhanam, James Gabriel, Peter Grasch, Oncel Tuzel, et~al.
\newblock Fastvlm: Efficient vision encoding for vision language models.
\newblock In \emph{Proceedings of the Computer Vision and Pattern Recognition Conference}, pages 19769--19780, 2025.

\bibitem[Vera et~al.(2025)Vera, Dua, Zhang, Salz, Mullins, Panyam, Smoot, Naim, Zou, Chen, et~al.]{vera2025embeddinggemma}
Henrique~Schechter Vera, Sahil Dua, Biao Zhang, Daniel Salz, Ryan Mullins, Sindhu~Raghuram Panyam, Sara Smoot, Iftekhar Naim, Joe Zou, Feiyang Chen, et~al.
\newblock Embeddinggemma: Powerful and lightweight text representations.
\newblock \emph{arXiv preprint arXiv:2509.20354}, 2025.

\bibitem[Wang et~al.(2024)Wang, Yuan, Zhang, and Sun]{wang2024tarsier}
Jiawei Wang, Liping Yuan, Yuchen Zhang, and Haomiao Sun.
\newblock Tarsier: Recipes for training and evaluating large video description models.
\newblock \emph{arXiv preprint arXiv:2407.00634}, 2024.

\bibitem[Wang and Isola(2020)]{wang2020understanding}
Tongzhou Wang and Phillip Isola.
\newblock Understanding contrastive representation learning through alignment and uniformity on the hypersphere.
\newblock In \emph{International conference on machine learning}, pages 9929--9939. PMLR, 2020.

\bibitem[Wei et~al.(2022)Wei, Wang, Schuurmans, Bosma, Xia, Chi, Le, Zhou, et~al.]{wei2023chainofthoughtpromptingelicitsreasoning}
Jason Wei, Xuezhi Wang, Dale Schuurmans, Maarten Bosma, Fei Xia, Ed~Chi, Quoc~V Le, Denny Zhou, et~al.
\newblock Chain-of-thought prompting elicits reasoning in large language models.
\newblock \emph{Advances in neural information processing systems}, 35:\penalty0 24824--24837, 2022.

\bibitem[Wu et~al.(2024)Wu, Chen, Lin, Wang, Gao, Xu, Xu, Hu, Chen, and Shou]{wu2024videollm}
Shiwei Wu, Joya Chen, Kevin~Qinghong Lin, Qimeng Wang, Yan Gao, Qianli Xu, Tong Xu, Yao Hu, Enhong Chen, and Mike~Zheng Shou.
\newblock Videollm-mod: Efficient video-language streaming with mixture-of-depths vision computation.
\newblock \emph{Advances in Neural Information Processing Systems}, 37:\penalty0 109922--109947, 2024.

\bibitem[Xu et~al.(2016)Xu, Mei, Yao, and Rui]{xu2016msr}
Jun Xu, Tao Mei, Ting Yao, and Yong Rui.
\newblock Msr-vtt: A large video description dataset for bridging video and language.
\newblock In \emph{Proceedings of the IEEE conference on computer vision and pattern recognition}, pages 5288--5296, 2016.

\bibitem[Yao et~al.(2024)Yao, Yu, Zhang, Wang, Cui, Zhu, Cai, Li, Zhao, He, et~al.]{yao2024minicpm}
Yuan Yao, Tianyu Yu, Ao~Zhang, Chongyi Wang, Junbo Cui, Hongji Zhu, Tianchi Cai, Haoyu Li, Weilin Zhao, Zhihui He, et~al.
\newblock Minicpm-v: A gpt-4v level mllm on your phone.
\newblock \emph{arXiv preprint arXiv:2408.01800}, 2024.

\bibitem[Zhai et~al.(2023)Zhai, Mustafa, Kolesnikov, and Beyer]{zhai2023sigmoid}
Xiaohua Zhai, Basil Mustafa, Alexander Kolesnikov, and Lucas Beyer.
\newblock Sigmoid loss for language image pre-training.
\newblock In \emph{Proceedings of the IEEE/CVF international conference on computer vision}, pages 11975--11986, 2023.

\bibitem[Zhou et~al.(2025)Zhou, Pan, LeCun, and Pinto]{zhou2025dino}
Gaoyue Zhou, Hengkai Pan, Yann LeCun, and Lerrel Pinto.
\newblock Dino-wm: World models on pre-trained visual features enable zero-shot planning.
\newblock In \emph{Forty-second International Conference on Machine Learning}, 2025.

\bibitem[Zhou et~al.(2018)Zhou, Xu, and Corso]{zhou2018towards}
Luowei Zhou, Chenliang Xu, and Jason Corso.
\newblock Towards automatic learning of procedures from web instructional videos.
\newblock In \emph{Proceedings of the AAAI conference on artificial intelligence}, volume~32, 2018.

\bibitem[Zhou et~al.(2024)Zhou, Arnab, Buch, Yan, Myers, Xiong, Nagrani, and Schmid]{zhou2024streaming}
Xingyi Zhou, Anurag Arnab, Shyamal Buch, Shen Yan, Austin Myers, Xuehan Xiong, Arsha Nagrani, and Cordelia Schmid.
\newblock Streaming dense video captioning.
\newblock In \emph{Proceedings of the IEEE/CVF Conference on Computer Vision and Pattern Recognition}, pages 18243--18252, 2024.

\bibitem[Zhukov et~al.(2019)Zhukov, Alayrac, Cinbis, Fouhey, Laptev, and Sivic]{zhukov2019cross}
Dimitri Zhukov, Jean-Baptiste Alayrac, Ramazan~Gokberk Cinbis, David Fouhey, Ivan Laptev, and Josef Sivic.
\newblock Cross-task weakly supervised learning from instructional videos.
\newblock In \emph{Proceedings of the IEEE/CVF Conference on Computer Vision and Pattern Recognition}, pages 3537--3545, 2019.

\end{thebibliography}
